\documentclass[11pt]{article}

\usepackage[final]{acl}
\usepackage{graphicx}
\usepackage{times}
\usepackage{latexsym}
\usepackage{amsmath} 
\usepackage{multirow}
\usepackage[table,xcdraw]{xcolor}
\usepackage[T1]{fontenc}
\usepackage{multirow}

\usepackage[utf8]{inputenc}

\usepackage{microtype}

\usepackage{inconsolata}
\usepackage{graphicx}
\usepackage{amsfonts}
\usepackage{colortbl}

\title{Activation-Efficient Fine-Tuning via Sequence-Compressed Low-Rank Subspaces}

\author{
 \textbf{Irene Tenison\textsuperscript{1}},
 \textbf{Stella Ahn\textsuperscript{1}},
 \textbf{Miriam Kim\textsuperscript{1,2}},
 \textbf{Ebtisam Alshehri\textsuperscript{1}},
 \textbf{Lalana Kagal\textsuperscript{1}}
\\
\\
 \textsuperscript{1}MIT CSAIL,
 \textsuperscript{2}Harvard SEAS
\\
}

\begin{document}
\maketitle
\begin{abstract}
Parameter-Efficient Fine-Tuning (PEFT) has become the standard approach for adapting large language models (LLMs). However, despite reducing trainable parameter counts, many PEFT methods remain memory-bound by token-level activation intermediates that scale linearly with sequence length, limiting their practicality for long-context and on-device adaptation. Unlike prior memory-efficient PEFT approaches that preserve token-level adapter activations, LARS (Low-memory Activation-Rank Subspace) reduces the sequence-length scaling of stored adapter activation intermediates from $\mathcal{O}(BSRL)$ to $\mathcal{O}(BRL)$ by performing adaptation within a sequence-level low-rank subspace. Through pooled feature extraction, low-rank subspace modulations, and residual projection, LARS reduces the overall \textit{memory footprint by 23\% on GPUs (Figure \ref{fig:memory_decomposition}) and 47.8\% on CPUs (Table \ref{tab:cpu}) relative to LoRA at $S=1024$, with relative savings increasing with context length}. Experiments across reasoning, understanding, long-context, retrieval, and token-level tasks show that \textit{LARS maintains competitive performance relative to LoRA} while substantially improving memory efficiency. Beyond GPUs, LARS also \textit{reduces average CPU power and energy consumption per optimization step by 22.43\% and 22.06\%, respectively}. These results demonstrate that activation-efficient adaptation provides a practical path toward resource-constrained and on-device fine-tuning of LLMs.
\end{abstract}

\section{Introduction}

LLMs and transformer-based architectures have become the backbone of modern natural language processing \cite{attentionneed,llama}. While these models exhibit remarkable zero-shot capabilities, fine-tuning remains essential for specialized tasks, privacy-preserving local adaptation, and low-latency personalization in mobile and edge environments \cite{neverstartfromscratch}. However, the high memory requirements of LLM adaptation pose a significant barrier to deployment on resource-constrained hardware, where available memory is often restricted to a few gigabytes.

To address these constraints, the research community has pivoted toward PEFT\cite{peftsurvey}. Methods such as LoRA \cite{lora}, prefix tuning \cite{prefixtuning}, and IA3 \cite{ia3} update only a minute fraction of the model's total weights—often by orders of magnitude—while maintaining competitive downstream accuracy. This success has solidified \textit{a fundamental design assumption: that reducing the number of trainable parameters directly translates to improved deployability in memory-limited environments.}

\begin{figure}[t]
\centering
\includegraphics[width=0.85\columnwidth, type=pdf, ext=.pdf, read=.pdf]{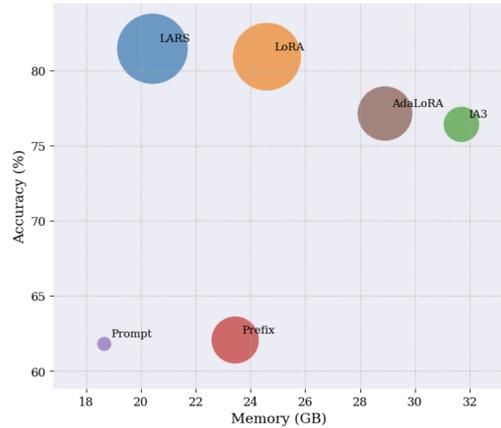}
\caption{Accuracy vs. peak memory (GB) for state-of-the-art PEFT methods. Bubble size represents the count of trainable parameters. Our analysis reveals a critical disconnect: trainable-parameter count is a poor proxy for actual memory footprint. }
\label{fig:accuracy-memory}
\end{figure}


As illustrated in Figure~\ref{fig:accuracy-memory}, there is a striking lack of correlation between a method's parameter efficiency and its actual physical memory footprint during adaptation. For instance, IA3 — one of the most parameter-efficient methods—requires significantly more peak memory than LoRA, despite having fewer trainable weights. 

\begin{figure}[t]
\centering
\includegraphics[width=\columnwidth]{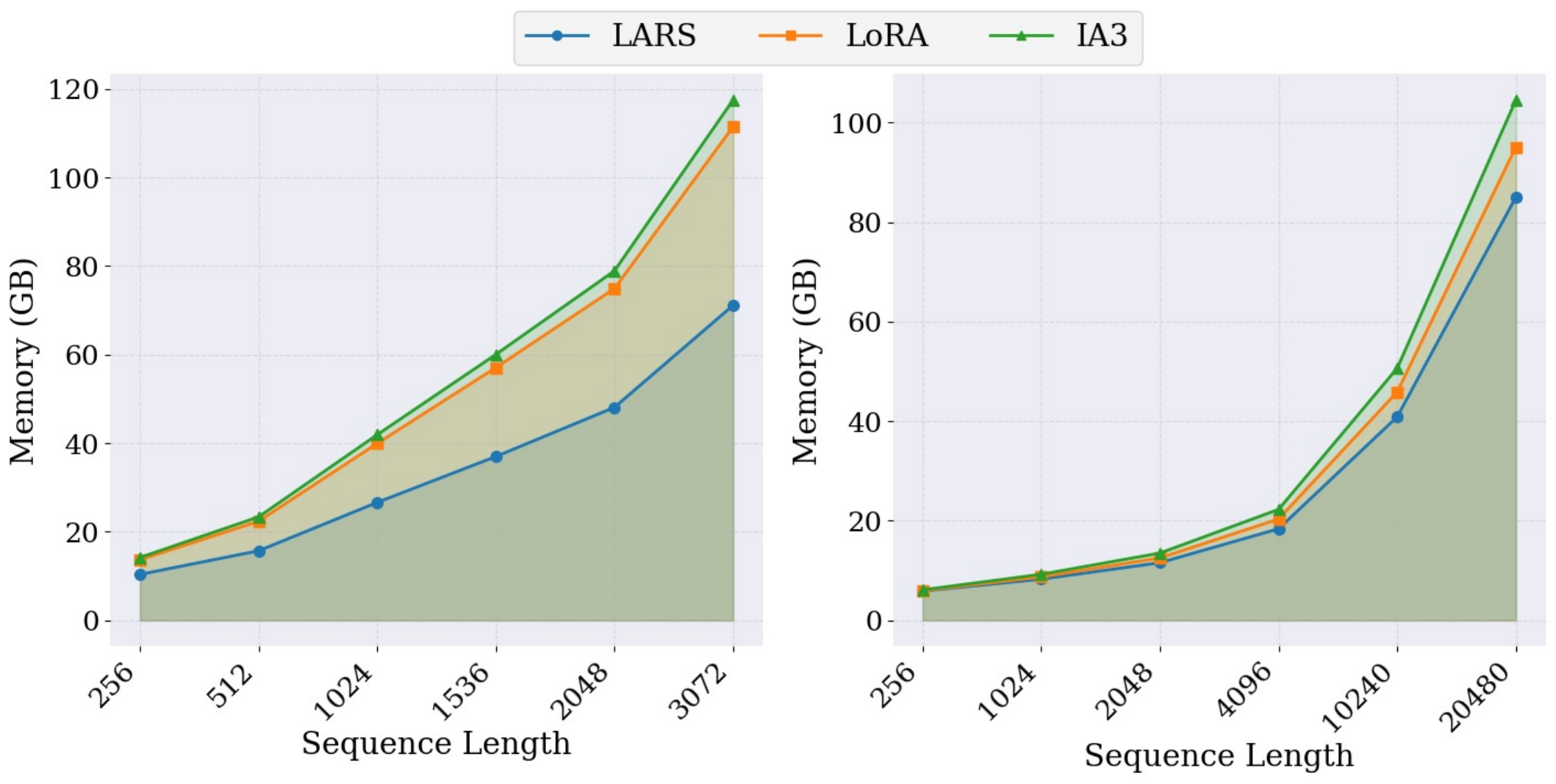}
\caption{Peak memory scaling vs. sequence length. (Left) Without GC and with GC enabled (Right), LARS defines a more efficient Pareto frontier, enabling the fine-tuning of sequences longer than LoRA within the same hardware constraints.}
\label{fig:s_dependence}
\end{figure}

Following prior works \cite{checkpointing, meft, lorafa}, we argue that the prevailing focus on parameter count overlooks the primary bottleneck of on-device adaptation - intermediate activation storage. Because most PEFT methods leave the forward computational graph largely unchanged, they still incur massive activation overhead that scales with batch size and sequence length, regardless of how few parameters are updated \cite{flashattention}. Consequently, reducing parameter count alone provides diminishing returns when activation memory dominates the peak footprint \cite{256kb}. 

\begin{figure}[t]
\centering
\includegraphics[width=0.9\columnwidth]{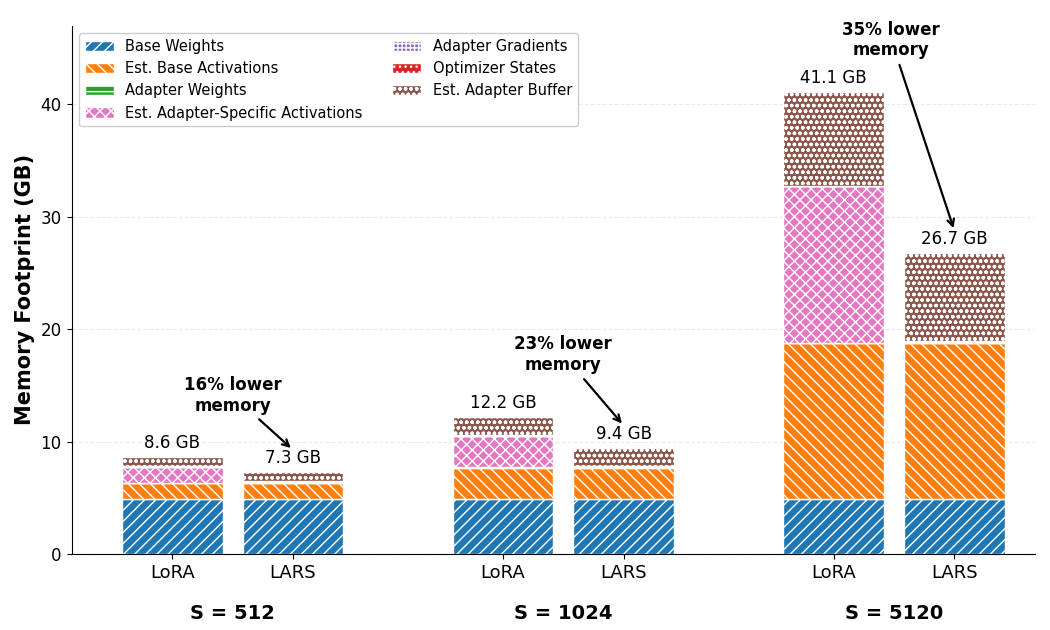}
\caption{Memory footprint breakdown of Llama 3.2 1B with LoRA and LARS across sequence lengths 512, 1024, and 5120. LARS reduces the sequence-length-dependent activation overhead of the adapter, yielding 12\%, 23\%, and 34\% lower overall memory consumption, respectively. Adapter-specific weights, gradients, and optimizer states are comparatively small and are not visually prominent. Estimated adapter-specific activations are the adapter internals that are reduced by LARS. Adapter buffer includes activations of adapter outputs and this scales with $S$. More details in Appendix \ref{app:memory}}
\label{fig:memory_decomposition}
\end{figure}

We introduce LARS (Low-memory Activation-Rank Subspace), an adaptation framework that trades token-level granularity for substantially lower adapter activation memory. U\textit{nlike prior memory-efficient PEFT methods that primarily reduce sparsity or backward-computation overhead \cite{dropbp, lisa, lorafa} while preserving token-level adapter activations, LARS reduces the sequence-length scaling of the adapter’s retained trainable intermediates themselves.} By operating within a sequence-level low-rank subspace, LARS lowers the hardware barrier for long-context and on-device adaptation while remaining complementary to existing systems-level memory optimizations. Our key contributions are:
\begin{itemize}
    \item  We propose LARS which performs adaptation in a sequence-pooled, low-rank subspace, reducing the size of stored adapter activations as illustrated in Figure \ref{fig:method}.
    \item LARS reduces the overall memory footprint by 23\% on GPUs\% on GPUs (Figure \ref{fig:memory_decomposition}) and 
$47.8\% $ on CPUs (Table \ref{tab:cpu}) in comparison to LoRA over $S=1024$. Furthermore, the relative memory savings increase with sequence length as shown in Figure \ref{fig:s_dependence}.

\item Our experiments across reasoning, understanding, long-context, and token-level tasks (Tables \ref{baselines}, \ref{fig:long-context1}, \ref{tab:token_level}) using different models further show that LARS maintains comparable accuracy to LoRA. 
    \item LARS reduces average power and energy consumption per step by 22.43\% and  22.06\% respectively in comparison to LoRA on CPUs (Table \ref{tab:cpu}).
\end{itemize}

\section{Background}
\label{background}
Adapting LLMs on-device requires understanding the runtime memory footprint beyond parameter counts. Peak memory is dominated by:
$
M_{peak} \approx M_{params}+M_{grads}+M_{opt}+M_{acts}
$
where $M_{params}, M_{grads}, M_{opt}, \text{ and } M_{acts}$ denote the memory consumed by model parameters, $\theta$, gradients of trainable parameters, optimizer states, and activations respectively.


Parameter-related terms scale with model size; $M_{params}=\mathcal{O}(|\theta|)$ and $M_{grads}, M_{opt}=\mathcal{O}(|\theta_{trainable}|)$. In contrast,activation memory depends on both model and input dimensions. For a transformer with depth L, hidden dimension H, batch size B, and sequence length S, $M_{acts}=\mathcal{O}(BSHL)$. This creates a fundamental asymmetry: while parameter memory is largely static, activation memory grows directly with sequence length $S$ and batch size $B$, often becoming the dominant bottleneck during adaptation.

\subsection{Memory Consumption in PEFT Methods}
PEFT methods operate in the regime where the trainable parameter fraction $\rho=\frac{|\theta_{trainable}|}{|\theta|} << 1$ \cite{lora,peftsurvey}. This substantially reduces gradient and optimizer-state memory. However, reducing trainable parameters alone does not significantly lower peak memory usage because activations remain dominant.



Most PEFT methods preserve the original token-level computational graph. For example, LoRA\cite{lora} applies a low-rank update of the form $Wx + ABx$, where $A \in \mathbb{R}^{H\times R}, B \in \mathbb{R}^{R\times H}, \text{ and } R<<H$. Although the trainable parameter count scales with R, the intermediate hidden states must still be retained for gradient computation. Consequently, activation memory scales as 
$M_{\text{acts}} = \underbrace{\mathcal{O}(BSHL)}_{\text{Base Activations}} + \underbrace{\mathcal{O}(BSRL)}_{\text{Adapter Activations}}
$.\\

As shown in Figure \ref{fig:s_dependence}, this token-level dependence causes memory usage to grow linearly with sequence length even under system-level optimizations such as gradient checkpointing. Existing PEFT approaches therefore remain constrained by a practical sequence-length ceiling on memory-limited hardware. 

Figure \ref{fig:memory_decomposition} and Table \ref{mem_decomp_table} further show that adapter weights, gradients, and optimizer states contribute only a negligible fraction of the overall memory footprint and peak memory during adaptation is dominated by activations. Although the base model has more parameters than the adapters or PEFT modules, it is frozen and does not retain as much activations for backpropagation. However, the activation memory of the base model remains substantial and scales with sequence length. Existing PEFT methods introduces additional trainable token-level intermediate tensors whose retained runtime memory grows with sequence length and has to be retained during backpropagation. Consequently, the relative contribution of adapter-specific activation memory becomes increasingly pronounced in long-context settings. These observations motivate a shift from parameter-efficient adaptation toward activation-efficient adaptation.

These observations motivate a shift from parameter-efficiency toward activation-efficiency. \textit{LARS addresses this bottleneck by decoupling adapter-specific activations from sequence length, asymptotically reducing adapter activation complexity from $O(BSRL)$ to $O(BRL)$.} Notably, LARS reduces overall memory by 12–34\% in comparison to LoRA, while its estimated adapter-specific activation memory remains minimal and effectively constant as sequence length \(S\) increases (Table \ref{mem_decomp_table}).



\section{Methods}



 


\paragraph{Problem Setting: } Let $M$ be a pre-trained Transformer with $L$ layers and hidden dimension $H$. For an input sequence $X\in\mathbb{R}^{B\times S\times H}$, standard PEFT methods (e.g., LoRA) compute an update $\Delta X = f(X;\theta_{trainable})$. The primary memory overhead in these methods arises from the adapter-specific activations that must be materialized and stored to compute the gradient $\nabla_{\theta_{trainable}}$. While base activations can be managed via GC, the adapter's intermediate tensors — of shape $[B,S,R]$ —still scale linearly with sequence length $S$. Our objective is to design an adaptation function $f(.)$ such that the adapter-specific activation memory is decoupled from $S$, reducing its complexity from $\mathcal{O}(BSRL)$ to $\mathcal{O}(BRL)$. 

\subsection{LARS (Low-memory Activation-Rank Subspace) Adaptation}

\begin{figure}[t]
\centering
\includegraphics[width=1\columnwidth]{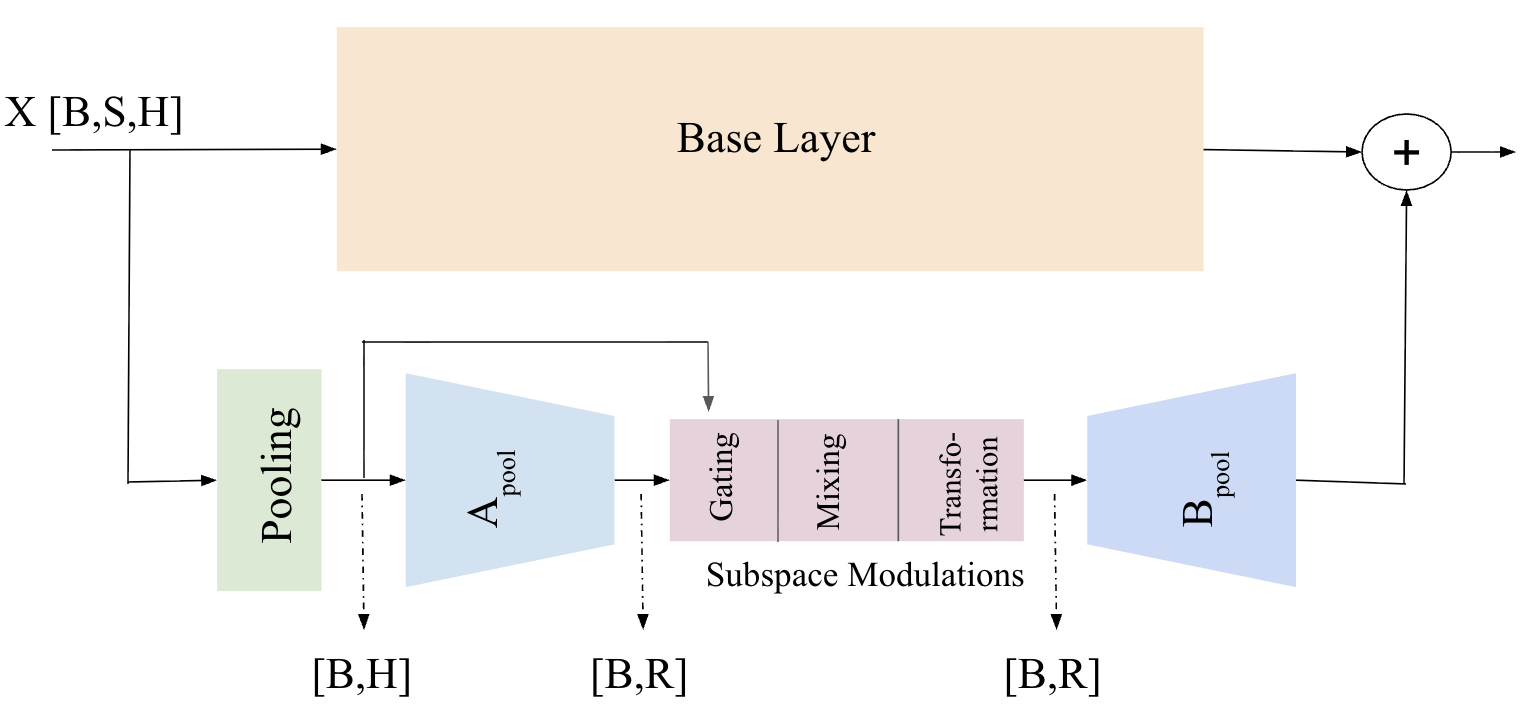}
\caption{Overview of LARS. Unlike LoRA, which performs token-level low-rank adaptation and retains sequence-dependent intermediates of size $\mathbb{R}^{B \times S \times R}$ during backpropagation, LARS compresses hidden states into a pooled sequence-level representation and performs adaptation entirely within a compact low-rank subspace $\mathbb{R}^{B \times R}$. As a result LARS reduces adapter activation scaling from $\mathcal{O}(BSRL)$ in LoRA to $\mathcal{O}(BRL)$.}
\label{fig:method}
\end{figure}

We propose LARS (Low-memory Activation-Rank Subspace) as illustrated in Figure \ref{fig:method}, an adaptation architecture designed to decouple the memory footprint of trainable modules in PEFT from the input sequence length $S$. While standard PEFT methods maintain a token-parallel intermediate activations per layer $[B,S,R]$, LARS executes adaptation within a compressed latent manifold, which ensures that the intermediate tensors required for gradient computation in the adapters are $S$-independent. The LARS execution pipeline consists of three stages: (i) Pooled Feature Extraction, which projects the sequence into a global context vector; (ii) Low-Rank Subspace Adaptation, which performs non-linear modulation in a rank-reduced space; and (iii) Residual Integration, which projects the learned updates back to the original manifold via broadcasting.

\subsubsection{Pooled Feature Extraction}
\label{pooling}
In the first stage of LARS, we address the activation bottleneck by projecting the sequence $X \in \mathbb{R}^{B \times S \times H}$ into a condensed global context vector $X_{pool} \in \mathbb{R}^{B \times H}$. This operation is the critical "memory-breaker" that enables LARS to avoid storing token-level hidden states for the adapter's backward pass. We propose two distinct strategies to manage the interplay between representation quality and hardware constraints.

1. \textbf{Heuristic-Driven Fixed Pooling} This strategy aggregates global semantic information without introducing new trainable parameters. We employ a hybrid mean-pooling scheme:

\begin{equation}
  x_{pool} = \frac{1}{S}\sum_{i=1}^S X_i +X_S   
\end{equation}

By augmenting the sequence mean with the final token representation $X_S$, we ensure the pooled vector captures both the global context and the recency bias inherent in causal transformers \cite{sink}. Fixed pooling is $\mathcal{O}(1)$ in terms of additional activation memory, making it the optimal choice for  device constraints.

2. \textbf{Context-Aware Learned Pooling} 
For tasks requiring finer semantic precision, we also propose Context-Aware Learned Pooling. This variant adaptively weights "high-information" tokens (e.g., verbs in reasoning tasks) over redundant padding. While this adds a marginal $\mathcal{O}(BH)$ activation term for attention weights, it remains an order of magnitude smaller than the $\mathcal{O}(BSH)$ requirement of token-level adapters \cite{pooling}. We designate fixed pooling as the default throughout this work to prioritize maximum memory efficiency for on-device constraints.




\subsubsection{Low-Rank Subspace Modulations}
\label{subspace}
Given the pooled representation $x_{pool} \in \mathbb{R}^{B\times H}$, LARS projects it into a low-rank subspace of dimension $R<<H$. 
\begin{equation}
    h=x_{pool}A_{pool}; \quad A_{pool} \in \mathbb{R}^{H \times R}
\end{equation}

While $A_{pool}$ is conceptually similar to the input projection in LoRA \cite{lora}, in LARS its goal is to map a global sequence summary into a feature space that remains informative. To maximize the representational capacity of this singular vector $h\in \mathbb{R}^{B \times R}$, we introduce a tiered modulation framework.

\paragraph{Feature-Space Gating} To ensure the model remains sensitive to different input contexts despite the sequence collapse, we apply an instance-conditioned gate $g \in \mathbb{R}^{B\times R}$. The gating signal is computed by integrating the global context with the local subspace features:
\begin{equation}
g = \sigma(\tau_1 \cdot W_x \cdot x_{\text{pool}} + \tau_2 \cdot W_h \cdot \textit{LN}(h))
\end{equation}
where $W_x \in \mathbb{R}^{H\times R}$ and $W_h \in \mathbb{R}^{R\times R}$ are learnable linear projections, $\textit{LN}(.)$ denote layer normalization, and $\sigma$ denotes the element-wise sigmoid function. We introduce learnable temperature scalars $\tau_1$ and $\tau_2$ to calibrate the relative influence of the global and local representations before the sigmoid activation. This mechanism allows the model to dynamically amplify or suppress specific subspace coordinates based on the input's global semantic signature. 

\begin{figure*}[h]
\centering
\includegraphics[width=2.1\columnwidth]{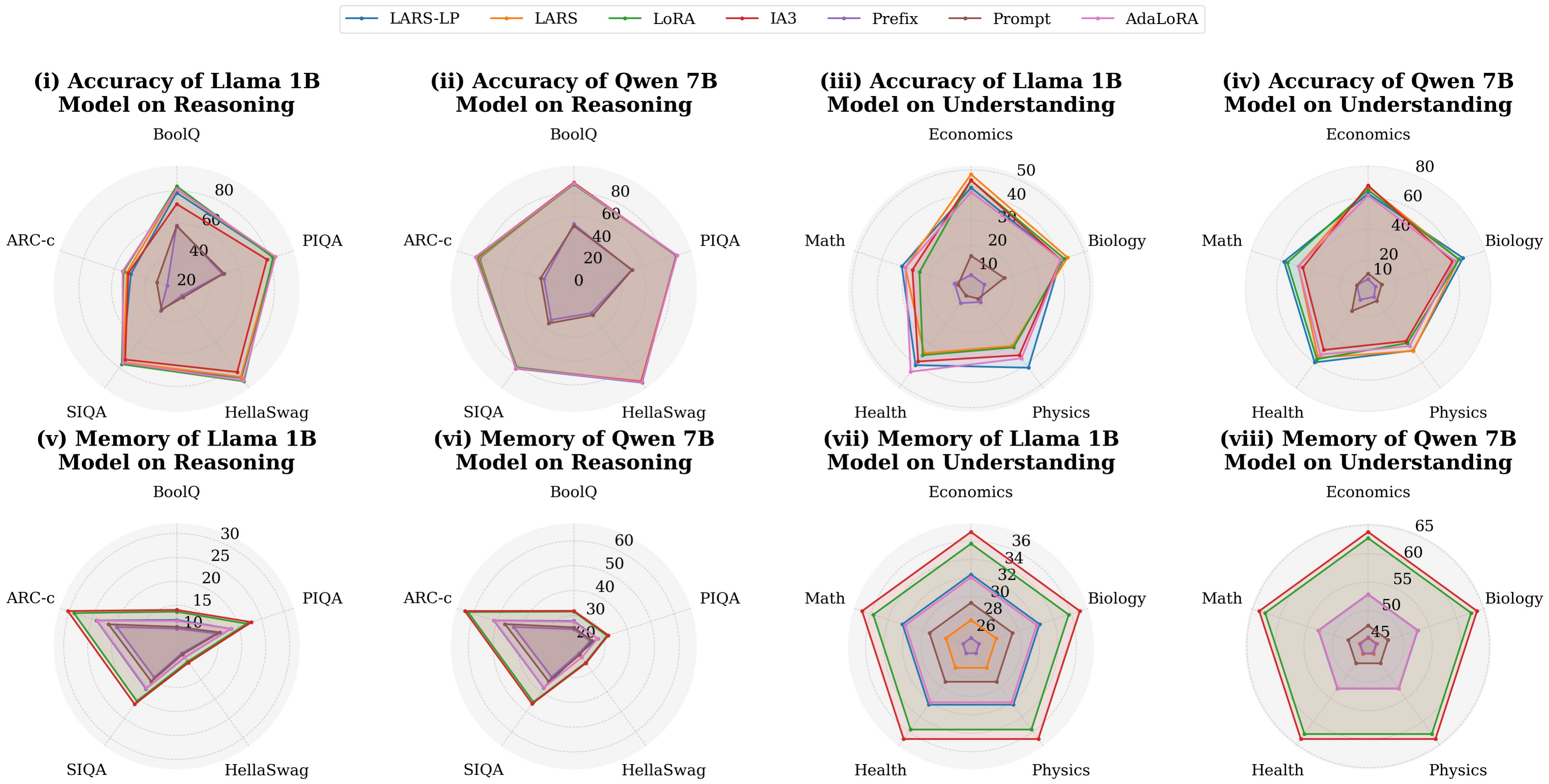}
\caption{Accuracy and Memory on Llama 1B and Qwen 7B on reasoning and understanding tasks across various datasets. While LARS consumes significantly lower memory than LoRA, IA3, and AdaLoRA, the performance is comparable on reasoning and understanding tasks.}
\label{fig:baselines}
\end{figure*}

\paragraph{Inter-Rank Mixing} Standard low-rank updates typically treat each subspace dimension as an independent component, limiting the model's capacity to capture feature correlations. To approximate the expressive power of a full-rank transformation, we introduce a learnable mixing transformation $M_{mix} \in \mathbb{R}^{R \times R}$ applied directly to the gating vector. This implements relational gating, where the importance of a specific subspace coordinate is conditioned on the global state of the gating manifold:
\begin{equation}
g_{mix} = g M_{mix}, \quad \text{and} \quad h' = g_{mix} \odot h
\end{equation}
This facilitates cross-rank communication and higher-order dependencies without expanding the activation footprint beyond $\mathcal{O}(BR)$, providing a significant boost to adaptation capacity on constrained devices.

\paragraph{Subspace Non-Linear Transformation}  To further enhance the representational capacity of the adaptation module, we introduce a lightweight non-linear bottleneck within GELU activation in the rank-reduced manifold. This operation provides the functional complexity necessary to model non-linear residual updates without reverting to expensive token-level computations. 


\begin{figure}[t]
\centering
\includegraphics[width=\columnwidth]{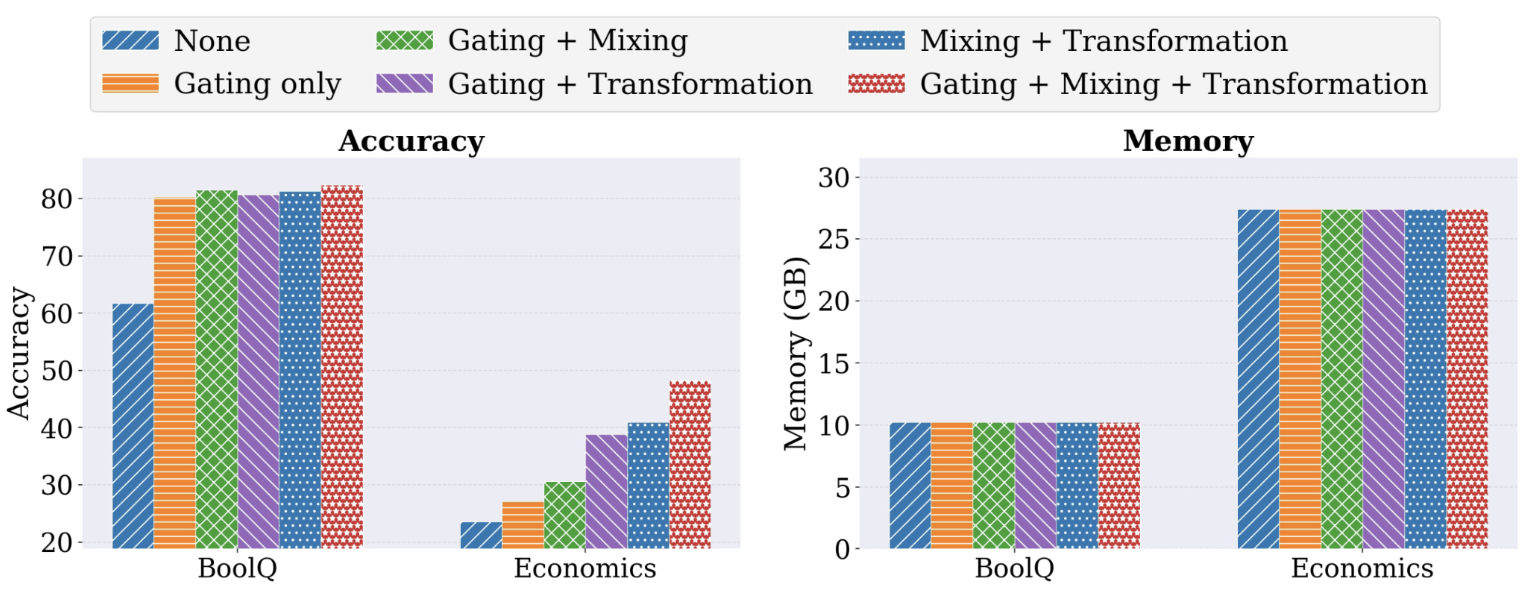}
\caption{Impact of Gating, Mixing, and Transformation components on Accuracy and Memory Usage (GB) across the BoolQ and Economics datasets.}
\label{fig:modulations}
\end{figure}

Figure \ref{fig:modulations} evaluates the contribution of the proposed subspace modulation components in LARS — Gating, Mixing, and Non-Linear Transformation. The results show that progressively combining these components consistently improves downstream accuracy, particularly on the more challenging Economics dataset, while introducing negligible additional memory overhead. This suggests that adaptation signals can be effectively modeled within a compact sequence-level low-rank manifold, where lightweight modulation operations recover representational expressivity without reintroducing expensive token-level activations. These findings further support the central design principle of LARS: preserving adaptation quality does not require maintaining full sequence-dependent adapter activations throughout training.

\begin{table*}[t]
\centering
\caption{ Average memory usage and accuracy of LARS and its learned pooling variant, LARS-LP on various reasoning, understanding, and long context tasks compared against baselines - LoRA, AdaLoRA, IA3, Prefix tuning, and Prompt Tuning. LARS consumes an average of $25\%$ lower memory than the baselines across tasks and models while performing comparably to LoRA.}
\label{baselines}
\resizebox{2\columnwidth}{!}{
\begin{tabular}{ccccccllccccll}
\hline
\rowcolor[HTML]{EFEFEF} 
                 & \textbf{(\%)}                                                & \multicolumn{6}{c}{\cellcolor[HTML]{EFEFEF}\textbf{Llama-3.2-1B}}                                                                                                                                                              & \multicolumn{6}{c}{\cellcolor[HTML]{EFEFEF}\textbf{Qwen2.5-7B-Instruct}}                                                                                                                                          \\
                 & \textbf{Trainable}                                           & \multicolumn{2}{c}{\textbf{Reasoning}} & \multicolumn{2}{c}{\textbf{Understanding}}                                                           & \multicolumn{2}{l}{\textbf{Long Context}}                                      & \multicolumn{2}{c}{\textbf{Reasoning}}                                            & \multicolumn{2}{c}{\textbf{Understanding}}                                        & \multicolumn{2}{l}{\textbf{Long Context}} \\
\rowcolor[HTML]{EFEFEF} 
                 & \multicolumn{1}{c|}{\cellcolor[HTML]{EFEFEF}\textbf{Params}} & \textbf{Mem.($\downarrow$)}    & \textbf{Acc.($\uparrow$)}   & \cellcolor[HTML]{EFEFEF}\textbf{Mem.($\downarrow$)}                    & \cellcolor[HTML]{EFEFEF}\textbf{Acc.($\uparrow$)} & \textbf{Mem.($\downarrow$)} & \multicolumn{1}{l|}{\cellcolor[HTML]{EFEFEF}\textbf{Acc.($\uparrow$)}} & \cellcolor[HTML]{EFEFEF}\textbf{Mem.($\downarrow$)} & \cellcolor[HTML]{EFEFEF}\textbf{Acc.($\uparrow$)} & \cellcolor[HTML]{EFEFEF}\textbf{Mem.($\downarrow$)} & \cellcolor[HTML]{EFEFEF}\textbf{Acc.($\uparrow$)} & \textbf{Mem.($\downarrow$)}     & \textbf{Acc.($\uparrow$)}     \\ \hline
\textbf{LoRA}    & \multicolumn{1}{c|}{0.45}                                    & 19.06              & \textbf{76.19}    & 35.72                                                      & 35.74                                   & 25.56           & \multicolumn{1}{l|}{57.91}                                   & 35.72                                   & 35.73                                   & 62.73                                   & 57                                      & 52.1                & 67.37               \\
\rowcolor[HTML]{EFEFEF} 
\textbf{AdaLoRA} & \multicolumn{1}{c|}{\cellcolor[HTML]{EFEFEF}0.45}            & 16.06              & 75.8              & 32.6                                                       & 37.4                                    & 23.44           & \multicolumn{1}{l|}{\cellcolor[HTML]{EFEFEF}\textbf{58.03}}  & 32.06                                   & 38.4                                    & 52.81                                   & 53.99                                   & 46.73               & \textbf{67.59}      \\
\textbf{IA3}     & \multicolumn{1}{c|}{0.01}                                    & 19.83              & 71.66             & 36.99                                 & 37.4                                    & 26.49           & \multicolumn{1}{l|}{54.13}                                   & 36.99                                   & 37.44                                   & 63.81                                   & 52.91                                   & 53                  & 62.02               \\
\rowcolor[HTML]{EFEFEF} 
\textbf{Prefix}  & \multicolumn{1}{c|}{\cellcolor[HTML]{EFEFEF}0.03}            & \textbf{13.47}     & 39.23             & \cellcolor[HTML]{EFEFEF}\textbf{25.49} & 9.05                                    & 22.30           & \multicolumn{1}{l|}{\cellcolor[HTML]{EFEFEF}24.62}           & \textbf{25.49}                          & 9.03                                    & \textbf{45.22}                          & 9.17                                    & 38.89               & 25.95               \\
\textbf{Prompt}  & \multicolumn{1}{c|}{0.003}                                   & 14.34              & 40.79             & 29.31                                  & 12.87                                   & 20.79           & \multicolumn{1}{l|}{24.39}                                   & 29.31                                   & 15.47                                   & 47.39                                   & 14.71                                   & 39.04               & 25.88               \\ \hline
\rowcolor[HTML]{EFEFEF} 
\textbf{LARS}    & \multicolumn{1}{c|}{\cellcolor[HTML]{EFEFEF}0.67}            & 13.55              & 75.6              & 27.44                                                      & 37.33                                   & \textbf{20.27}  & \multicolumn{1}{l|}{\cellcolor[HTML]{EFEFEF}57.18}           & 27.44                                   & 37.33                                   & 45.38                                   & 56.97                                   & \textbf{38.33}      & 67.18               \\
\textbf{LARS-LP} & \multicolumn{1}{c|}{0.67}                                    & 16.07              & 75.97             & 32.39                                                      & \textbf{39.32}                          & 23.19           & \multicolumn{1}{l|}{54.52}                                   & 32.39                                   & \textbf{39.32}                          & 52.87                                   & \textbf{59.19}                          & 44.28               & 63.9                \\ \hline
\end{tabular}}
\end{table*}

\subsubsection{Residual Projection and Integration } 
The final stage of LARS maps the adapted subspace features back to the original model manifold via broadcasting across $S$. The modulated representation $h'\in \mathbb{R}^{B\times R}$ is projected to the hidden dimension $H$ and integrated into the frozen backbone via a gated residual connection with learnable scalar $\alpha$ to calibrate the adapter's influence on the frozen features.
\begin{equation}
\begin{aligned}
    \text{out}[b,s,:] = \text{Base}_{\text{out}}&[b,s,:] + \alpha . \text{Adapter}_{\text{out}}[b,:]\\
    \text{Adapter}_{\text{out}} &= B_{pool}(h') \\
\end{aligned}
\end{equation}

The residual formulation \cite{petl} ensures the numerical stability of the pre-trained weights and mitigate catastrophic forgetting, as the adapter shifts the base model's distribution.

\paragraph{Summary } By collapsing the sequence dimension before the high-rank backward pass, LARS decouples the adapter’s activation memory footprint from input length, enabling scalable fine-tuning of longer contexts on memory-constrained devices. Combined with gating, inter-rank mixing, and non-linear transformations, LARS recovers the expressive capacity lost during pooling.  More details on the implementation including initializations and code snippets are given in Appendix \ref{code}. 

\section{Experiments and Results}
\subsection{Setup}
\paragraph{Datasets and Tasks} We evaluate LARS primarily across three task categories: (1) Commonsense Reasoning using a subset of the LLM-adapters benchmark \cite{reasoning} (BoolQ, PIQA, SIQA, HellaSwag, and ARC-c \cite{boolq, piqa, siqa, hellaswag, arc}); (2) General Understanding via five MMLU-Pro subjects \cite{mmlupro} (Economics, Biology, Physics, Health, and Math); and (3) Long-Context Comprehension using QuALITY \cite{quality} and RACE \cite{race}. These tasks measure semantic reasoning and document-level understanding under memory constraints. Additionally, to further understand the performance of LARS, which compresses token-level information in the adapters, we experiment on representative token-level and generation tasks - sequence labeling on CoNLL \cite{conll}, extractive QA on SQuAD \cite{squad}, and summarization on SAMSum \cite{samsum}. All theses datasets are public and easy to access.

\begin{figure}[t]
\centering
\includegraphics[width=0.85\columnwidth]{Images/Qwen_long.pdf}
\caption{Memory consumption and accuracy of Qwen 7B on long context Quality and RACE datasets}
\label{fig:long-context1}
\end{figure}

\paragraph{Baselines} We evaluate LARS against five standard PEFT methods: LoRA \cite{lora}, its rank-adaptive variant AdaLoRA \cite{adalora}, and the multiplicative scaling method IA3 \cite{ia3}. We also compare against the prompt-based approaches of Prefix \cite{prefixtuning} and Prompt Tuning \cite{prompttuning}. 


\paragraph{Implementation} Our experiments involve publicly available Llama-3.2-1B model that was run on NVIDIA L4OS GPU and Qwen2.5-7B-Instruct model that was run on H200 GPU as the primary base models for all reasoning, understanding, and long-context tasks. All experiments report results from a single run as the primary contribution of this work—reducing adapter-specific activation memory scaling—is determined by the architectural structure of the method and is largely invariant to initialization or random seed choice. The reported accuracy trends are comparative in nature. For most experiments we use LARS with fixed pooling as this paper focuses on a memory-constrained on-device setting. All baseline implementations are from Huggingface PEFT library with all default parameters. Appendix \ref{App:setup} gives more details on the implementation of LARS, datasets, hardware, hyperparameters, tuning, and memory and power measurement methodology used for various experiments. 

\subsection{Results and Discussion}
\label{results}
\subsubsection{Reasoning and Understanding} 
\label{main_results}
As shown in Table 1, LARS demonstrates a superior memory-to-accuracy trade-off compared to other baselines. The primary advantage of LARS lies in its significant reduction of the memory footprint while maintaining comparable performance. On the Qwen 7B and Llama3.2 1B models for understanding tasks, LARS requires only approximately 25\% lower memory on average than LoRA at $S=1024$. Despite this smaller footprint, LARS maintains competitive accuracy compared to LoRA. These results can be further validated with the dataset-specific and visual results in Figure \ref{fig:baselines}. While the memory-usage (bottom row) is visibly minimized for LARS, the accuracy contours (top row) remain largely overlapping with the most baselines. \textit{These findings suggest that LARS provides a favourable solution for deploying LLMs on hardware with limited VRAM, offering a more efficient path for fine-tuning without the typical performance degradation associated with aggressive memory optimization.}

\textbf{LARS-LP}, which is a variant of LARS with the learned pooling strategy introduced in Section \ref{pooling} instead of fixed pooling variant, successfully bridges the slight accuracy gap in LARS. In the Qwen 7B Reasoning task, LARS-LP improves the accuracy to 39.32\%, outperforming standard LARS (37.33\%) and other baselines although it consumes slightly more memory. \textit{This indicates that the learned pooling mechanism effectively retains critical features necessary for complex reasoning and understanding while maintaining the efficiency gains of the LARS architecture.}

\paragraph{Long Context:} For long-context datasets, across both models as shown in Figures \ref{fig:long-context1} and \ref{fig::long-context2} and datasets, LARS consistently demonstrates superior memory efficiency and comparable performance relative to all other high-performing adapters similar to reasoning and understanding tasks. Collectively, these results indicate that \textit{LARS provides a favourable trade-off for long-context applications, offering the high-rank representation power of LoRA with a memory profile more closely resembling (or bettering) more restrictive PEFT methods.}

\paragraph{Token-Level Fidelity: } 
\begin{table}[]
\centering
\caption{Additional experiments on representative token-level and generation tasks: sequence labeling (SL) and structured decoding (SD) on CoNLL, Extractive QA on SQuAD, and Summarization on SAMSum, to better understand the effect of sequences-compression in LARS.}
\label{tab:token_level}
\resizebox{1\columnwidth}{!}{
\begin{tabular}{llllllllllll}
\hline
\rowcolor[HTML]{EFEFEF} 
\multicolumn{1}{c}{\cellcolor[HTML]{EFEFEF}\textbf{}} &  & 
\multicolumn{2}{c}{\cellcolor[HTML]{EFEFEF}\textbf{CoNLL(SL)}} &  & 
\multicolumn{1}{c}{\cellcolor[HTML]{EFEFEF}\textbf{CoNLL(SD)}} &  & 
\multicolumn{2}{c}{\cellcolor[HTML]{EFEFEF}\textbf{SQuAD}} &  & 
\multicolumn{2}{c}{\cellcolor[HTML]{EFEFEF}\textbf{SAMSum}} \\
\multicolumn{1}{c}{\textbf{}} &  & 
\multicolumn{1}{c}{\textbf{acc}} & 
\multicolumn{1}{c}{\textbf{F1}} &  & 
\multicolumn{1}{c}{\textbf{Entity-F1}} &  & 
\multicolumn{1}{c}{\textbf{EM}} & 
\multicolumn{1}{c}{\textbf{F1}} &  & 
\multicolumn{1}{c}{\textbf{Rouge-1}} & 
\multicolumn{1}{c}{\textbf{Rouge-L}} \\ \hline

\rowcolor[HTML]{EFEFEF} 
LARS &  & 
\textit{\textbf{98.12}} & 
\textit{\textbf{0.78}} &  & 
\multicolumn{1}{c}{0.77} &  & 
0.556 & 0.644 &  & 0.277 & 0.204 \\

LoRA &  & 
96.11 & 0.73 &  & 
\multicolumn{1}{c}{\textit{\textbf{0.82}}} &  & 
\textit{\textbf{0.558}} & 
\textit{\textbf{0.646}} &  & 
\textit{\textbf{0.302}} & 
\textit{\textbf{0.230}} \\

\rowcolor[HTML]{EFEFEF} 
IA3 &  & 
96.89 & 0.75 &  & 
\multicolumn{1}{c}{0.73} &  & 
0.414 & 0.495 &  & 0.294 & 0.220 \\

AdaLoRA &  & 
96.68 & 0.76 &  & 
\multicolumn{1}{c}{0.75} &  & 
0.556 & 0.642 &  & 0.293 & 0.224 \\ \hline
\end{tabular}}
\end{table}

Despite operating in a sequence-pooled subspace, LARS remains competitive with standard PEFT methods across all tasks. As shown in Table \ref{tab:token_level}, on CoNLL, LARS achieves the highest accuracy and F1 score, suggesting that compressed sequence representations can still preserve strong token-level semantic information. On SQuAD, LARS performs comparably to LoRA and AdaLoRA, with only a marginal drop in EM and F1. For summarization on SAMSum, LoRA achieves the strongest ROUGE scores, while LARS remains comparable despite its substantially more compressed activation structure. We additionally conducted Needle-In-A-Haystack (NIAH) style passkey retrieval experiments and observed that LARS, LoRA, AdaLoRA, and IA3 achieved near-perfect accuracy at lengths up to 16k. However, performance across all methods declined at the 32k context length as shown in Table \ref{tab:niah}. Overall, \textit{these results indicate that LARS maintains comparable downstream performance even on tasks that traditionally benefit from token-level adaptation.}

\begin{figure}[t]
\centering
\includegraphics[width=\columnwidth]{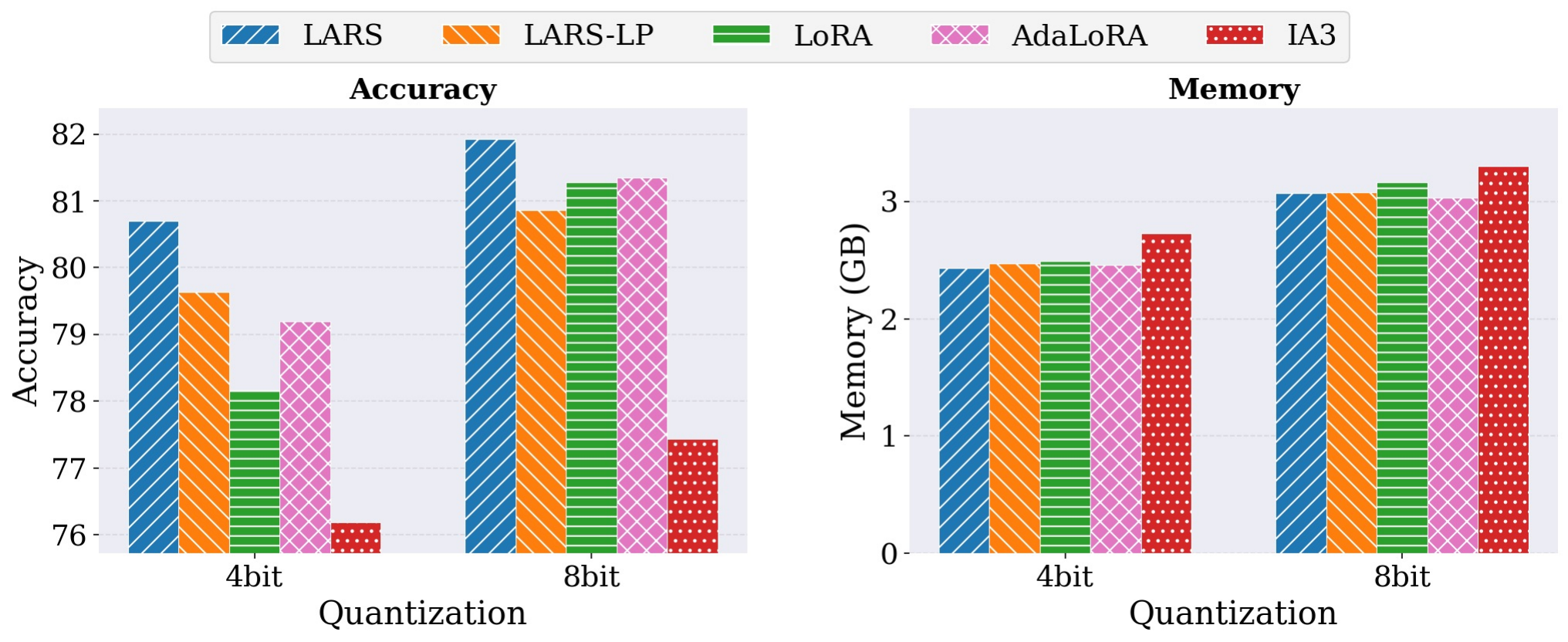}
\caption{Memory Usage (GB) and Accuracy of LARS and baselines with 4 bit and 8 bit quantization.}
\label{fig:quantization}
\end{figure}

\subsubsection{Effect of Quantization}
Figure \ref{fig:quantization} evaluates the effects of 4-bit versus 8-bit quantization on accuracy and memory usage across LARS and baselines. The plots show that LARS consistently achieves the highest performance in both 4 bit and 8 bit settings while consuming lower memory that baselines.
Similar characters were observed with other systems-level optimization methods like FlashAttention \cite{flashattention}, Checkpointing \cite{checkpointing}, as well as memory optimization methods like DropBP\cite{dropbp} and LoRA-FA\cite{lorafa} as shown in Appendix \ref{complementary methods}. Notably these methods are complementary to LARS and they can be combined to achieve better resource efficiency.

\begin{figure}[t]
\centering
\includegraphics[width=1\columnwidth]{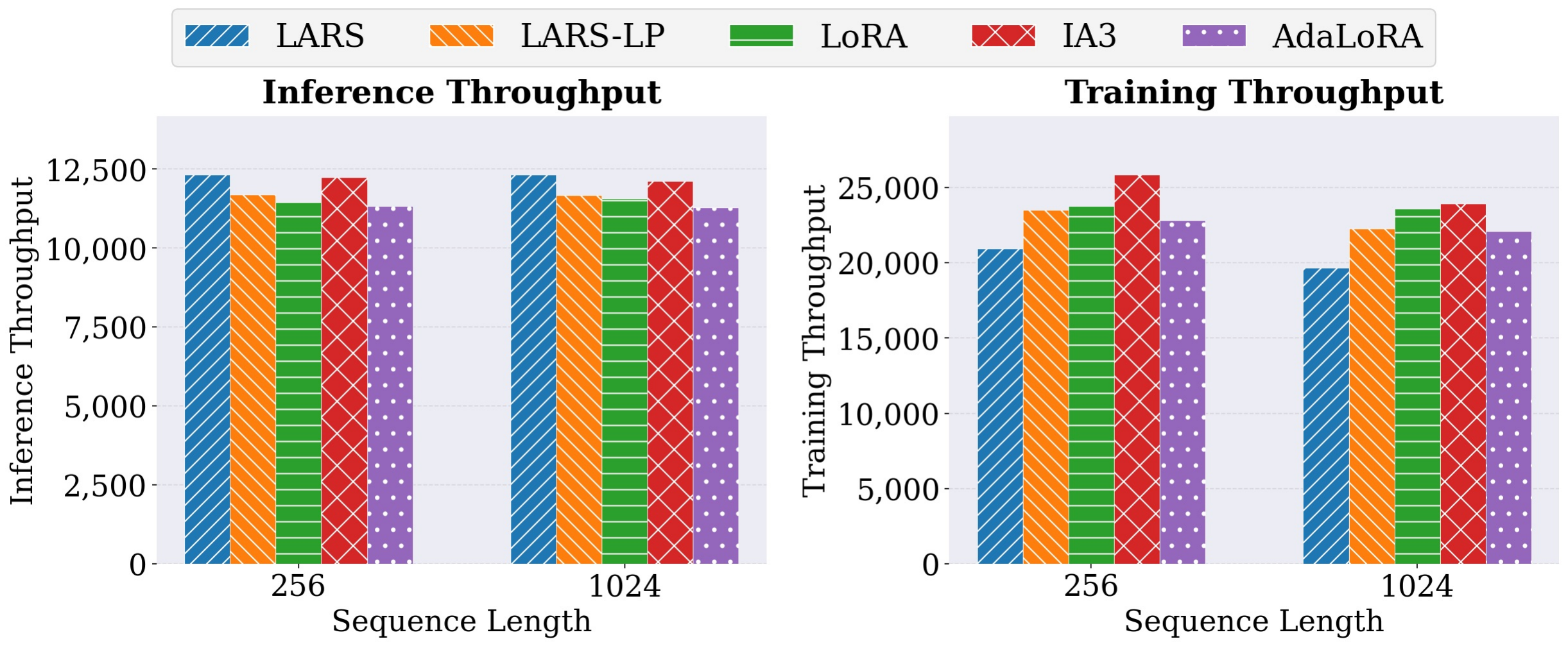}
\caption{ Throughput of LARS and other baselines during inference and fine-tuning across sequence lengths 256 and 1024.}
\label{fig:throughput}
\end{figure}
\subsubsection{Throughput Analysis}
\label{throughput_results}
Figure \ref{fig:throughput} illustrates the training and inference throughput across adaptation methods. While LARS is highly efficient for deployment in inference throughput and latency (see Figure \ref{fig:latency}, Appendix)—a divergence appears during fine-tuning. We observe that LARS is slightly slower in terms of training tokens per second compared to LoRA. However, this marginal reduction in speed is offset by a substantial gain in hardware accessibility and memory efficiency. This demonstrates that \textit{LARS effectively trades a fraction of computational throughput to mitigate the activation bottleneck} enabling long-context adaptation on devices where standard methods often fail.

\begin{table}[t]
\centering
\caption{Accuracy, memory consumption, throughput, average time per step, average power per step and average energy per step of LARS, LoRA and IA3 on AMD EPYC CPU. }
\label{tab:cpu}
\resizebox{1\columnwidth}{!}{

\begin{tabular}{lcccccc}
\hline
\rowcolor[HTML]{EFEFEF} 
\multicolumn{1}{c}{\cellcolor[HTML]{EFEFEF}} & \textbf{\begin{tabular}[c]{@{}c@{}}Accuracy\\ (\%)\end{tabular}} & \textbf{\begin{tabular}[c]{@{}c@{}}Memory\\ (GB)\end{tabular}} & \textbf{\begin{tabular}[c]{@{}c@{}}Throughput\\ (tok/sec)\end{tabular}} & \textbf{\begin{tabular}[c]{@{}c@{}}Time\\ (sec)\end{tabular}} & \textbf{\begin{tabular}[c]{@{}c@{}}Power\\ (W)\end{tabular}} & \textbf{\begin{tabular}[c]{@{}c@{}}Energy\\ (J)\end{tabular}} \\ \hline
\multicolumn{7}{c}{\textbf{S = 256}}                                                                                                                                                                                                                                                                                                                                                                                                                          \\ \hline
\rowcolor[HTML]{EFEFEF} 
\textbf{LARS}                                & 76.02                                                            & \textbf{15.824}                                                & 113.22                                                                  & 31.67                                                         & 74.64                                                        & 2238.16                                                       \\
\textbf{LoRA}                                & 77.56                                                            & 27.059                                                         & 153.25                                                                  & 32.15                                                         & 135.96                                                       & 4231.88                                                       \\
\rowcolor[HTML]{EFEFEF} 
\textbf{IA3}                                 & 67.5                                                             & 18.937                                                         & 101.78                                                                  & 27.8                                                          & 89.5                                                         & 2466.48                                                       \\ \hline
\multicolumn{7}{c}{\textbf{S = 512}}                                                                                                                                                                                                                                                                                                                                                                                                                          \\ \hline
\rowcolor[HTML]{EFEFEF} 
\textbf{LARS}                                & 74.58                                                            & \textbf{15.926}                                                & 111.88                                                                  & 30.65                                                         & 140.81                                                       & 4731.87                                                       \\
\textbf{LoRA}                                & 75.39                                                            & 28.164                                                         & 189.21                                                                  & 33.93                                                         & 160.86                                                       & 5071.31                                                       \\
\rowcolor[HTML]{EFEFEF} 
\textbf{IA3}                                 & 70.71                                                            & 19.077                                                         & 101.95                                                                  & 30.68                                                         & 161.95                                                       & 5731.88                                                       \\ \hline
\multicolumn{7}{c}{\textbf{S = 1024}}                                                                                                                                                                                                                                                                                                                                                                                                                         \\ \hline
\rowcolor[HTML]{EFEFEF} 
\textbf{LARS}                                & 78.24                                                            & \textbf{16.866}                                                & 111.56                                                                  & 30.24                                                         & 148.19                                                       & 4737.42                                                       \\
\textbf{LoRA}                                & 78.85                                                            & 32.314                                                         & 174.38                                                                  & 34.37                                                         & 164.14                                                       & 5405.82                                                       \\
\rowcolor[HTML]{EFEFEF} 
\textbf{IA3}                                 & 70                                                               & 19.827                                                         & 97.65                                                                   & 31.07                                                         & 183.96                                                       & 6556.87                                                       \\ \hline
\end{tabular}}
\end{table}

\subsubsection{Beyond GPUs}
\label{CPU_results}
To evaluate the practical efficiency and on-device deployment of the proposed LARS method, we compared the convergence accuracy, memory consumption, throughput, average time per step, average power per step and average energy per step of LARS against LoRA and IA3 on both CPUs. Across all settings, LARS consistently achieves the lowest memory consumption while maintaining competitive accuracy and throughput. At $S=1024$, LARS reduces peak memory usage by nearly 47.8\% relative to LoRA. LARS also reduces average power consumption and average energy per step across all sequence lengths by 22.43\% and 22.06\% compared to LoRA. These results demonstrate that \textit{LARS provides a favorable trade-off between adaptation quality and hardware resource usage, making it particularly suitable for memory- and energy-constrained CPU environments.}


\section{Related Works}

\paragraph{Parameter-Efficient Adaptation}
PEFT methods aim to adapt LLMs by updating a minimal fraction of the total weights. Methods like IA3\cite{ia3} and TENT\cite{tent} moved away from the adapters to achieve better efficiency with a performance drop. Most methods \cite{longlora,lora,adalora} remain bound by the "Activation Wall" — where $M_{acts}$ scales linearly with sequence length \cite{256kb,tinytrain,adabet}.

\paragraph{System-Level Memory Optimization} To mitigate activation overhead, the systems community has introduced GC\cite{checkpointing}, which trades FLOPs for memory by recomputing activations, and FlashAttention\cite{flashattention}, which optimizes the attention matrix. More recently, LISA\cite{lisa} reduce memory by freezing specific layers during the backward pass. However, these are optimizations do not alter the scaling of hidden-state tensors. LARS complements these approaches by decoupling adapter-specific activations from the sequence dimension.

\paragraph{On-Device \& Edge Intelligence} Deploying LLMs on resource-constrained hardware necessitates aggressive compression, typically through quantization (e.g., QLoRA \cite{qlora}) or Pruning \cite{tinytrain,adabet}. While these target $M_{params}$, LARS addresses the dynamic peak memory bottleneck. LARS approach aligns with the growing need for "tiny training" \cite{256kb} and local adaptation where RAM is often the limiting factor \cite{on-device_survey}.

\section{Conclusion}
We introduced an activation-efficient adaptation framework that reduces the sequence-length scaling of adapter activations by operating within a sequence-level low-rank subspace. By collapsing the sequence dimension prior to low-rank adaptation and residual reconstruction, LARS reduces the memory footprint associated with adapter-specific activation intermediates while maintaining competitive downstream performance across reasoning, understanding, long-context, retrieval, and on-device adaptation tasks. Across multiple models and hardware settings, LARS consistently improves the memory-to-accuracy trade-off relative to existing PEFT methods, while remaining complementary to system-level optimizations. Our results highlight the importance of evaluating activation behavior, rather than parameter count alone, when designing efficient adaptation methods for resource-constrained LLM training.

\section{Limitations}
LARS achieves its memory efficiency by compressing sequence-level representations prior to adaptation, which can reduce fine-grained token fidelity in tasks requiring precise lexical reconstruction or generation. While LARS remains competitive across sequence labeling, extractive QA, long-context retrieval, and reasoning tasks, modest degradations can appear on generation-oriented metrics such as BLEU, especially at extremely long sequences. Additionally, LARS introduces moderately more trainable parameters than some PEFT baselines at equivalent ranks and has currently been evaluated only on models up to 8B parameters. Future work includes extending activation-efficient adaptation to larger-scale models, improving token-level expressivity through alternative sequence modeling mechanisms, and exploring hybrid architectures that better preserve local token interactions while maintaining favorable activation scaling properties.

\section{Ethical Considerations}
During the preparation of this manuscript, we used ChatGPT (GPT-4) for assistance with manuscript organization, text editing and phrasing, and debugging; all technical ideas, implementations, experiments, and analyses were developed by the authors. By reducing the hardware and memory requirements of adaptation, LARS can improve accessibility to fine-tuning on resource-constrained devices and lower the energy cost associated with experimentation. However, improved accessibility may also enable adaptation of language models on harmful, biased, or privacy-sensitive datasets. Furthermore, the evaluated models inherit biases and limitations present in their pretraining corpora and downstream benchmarks. We encourage responsible deployment practices, transparent reporting of fine-tuning data, and careful consideration of safety and misuse risks when adapting LLMs in low-resource settings.

\bibliography{acl_latex}

\newpage

\appendix

\section{Background and Motivation}

\subsection{ The Fallacy of Parameter Count as a Memory Proxy}

\begin{figure}[h]
\centering
\includegraphics[width=0.85\columnwidth]{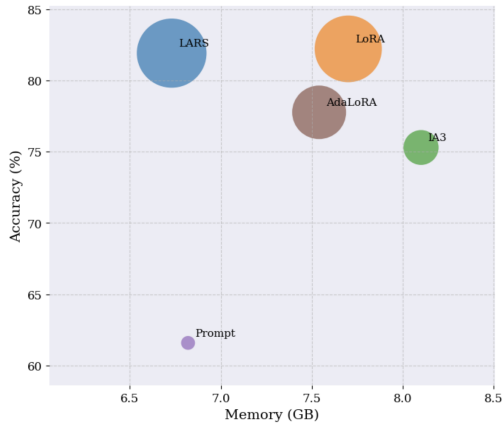}
\caption{Accuracy vs. peak training memory (GB) for state-of-the-art PEFT methods with CP. Even with checkpointing, the disconnect remains: trainable-parameter count is a poor proxy for actual memory footprint. }
\label{fig:accuracy-memory-cp}
\end{figure}

\begin{figure}[t]
\centering
\includegraphics[width=\columnwidth]{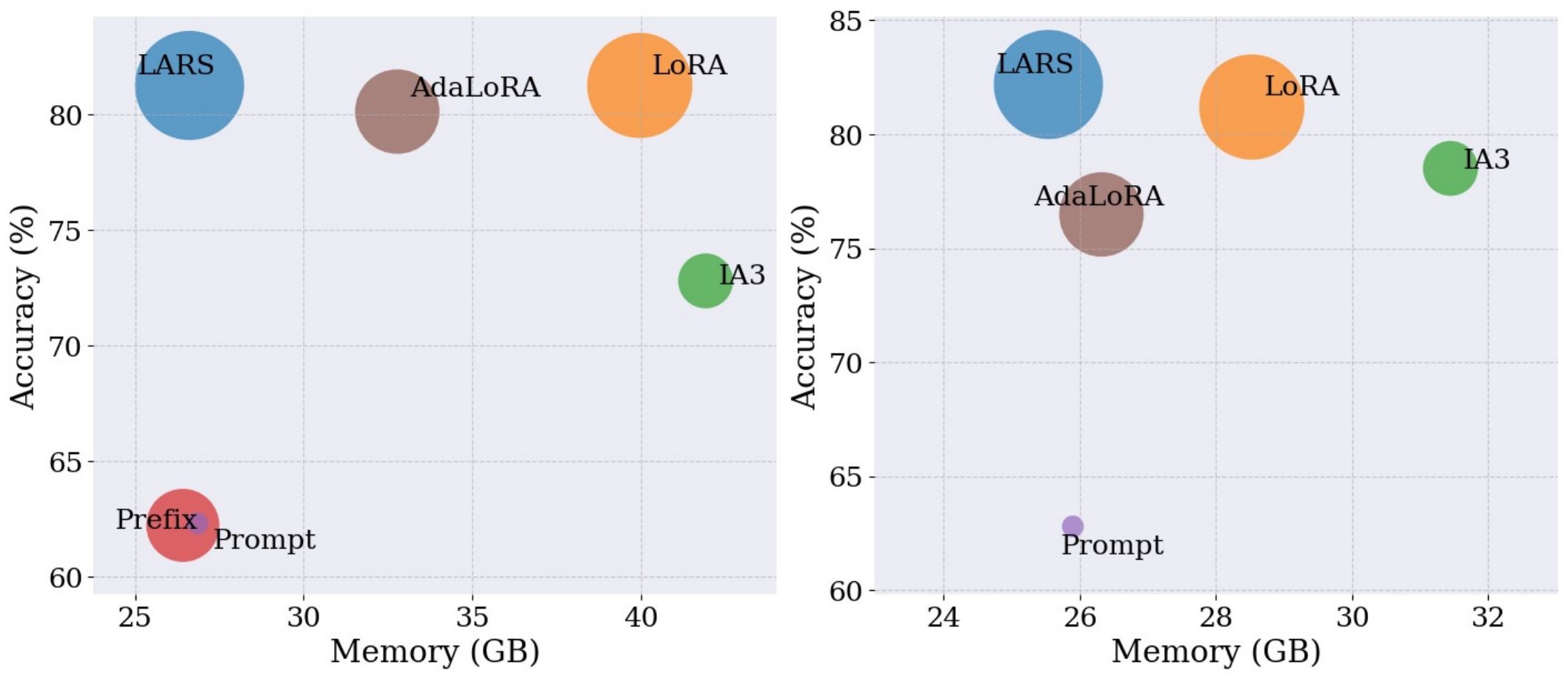}
\caption{Accuracy vs. peak training memory (GB) for state-of-the-art PEFT methods without (left) with (right) CP. Unlike Figures \ref{fig:accuracy-memory} and \ref{fig:accuracy-memory-cp}, these are based on static padding and the disconnect remains.}
\label{fig:accuracy-memory-static}
\end{figure}

While the PEFT literature traditionally uses the number of trainable parameters, $|\theta_{trainable}|$, as the primary metric for efficiency, our work demonstrates that this is a misleading proxy for actual on-device deployability. This subsection provides a detailed analysis of the empirical results presented in Figures 1, 3, and 4 that empirically validates this claim.

In Figure \ref{fig:accuracy-memory}, we plot Accuracy vs. Peak Training Memory for several state-of-the-art PEFT methods. The bubble size represents the relative count of trainable parameters. A "parameter-efficient" world would show a clear correlation where smaller bubbles (fewer parameters) appear on the left (lower memory). Instead, we observe a stochastic distribution. IA3 \cite{ia3} is among the most parameter-efficient methods ($\rho\approx0.003\%$), yet it occupies the highest peak memory region (32 GB). IA3 achieves parameter efficiency by learning multiplicative vectors that scale activations across the entire hidden dimension $H$ for every token $S$. Because these scaling factors must be applied to the full activation tensors at multiple points in the forward pass, the GPU must preserve these massive tensors for gradient computation, regardless of how few parameters are actually being updated.

Figure \ref{fig:accuracy-memory-cp} repeats this analysis with GC enabled. While GC is the standard "fix" for memory issues, the figure reveals that the disconnect between parameter count and memory footprint remains unchanged. GC reduces the constant factor of memory by recomputing activations during the backward pass. However, as Figure \ref{fig:accuracy-memory-cp} shows, the relative ordering of methods (e.g., LARS being the most efficient and IA3 being the least) remains the same. This shows that system-level tricks like GC shift the baseline but do not solve the structural flaw of token-level adaptation. Even with GC, a "small" method like LoRA can still be "large" in terms of relative peak memory.

One might argue that the memory variability in Figures \ref{fig:accuracy-memory} and \ref{fig:accuracy-memory-cp} is an artifact of dynamic sequence lengths or varying batch sizes. Figure \ref{fig:accuracy-memory-static} further supports this observation by showing that the disconnect persists under static padding, where sequence length is fixed. Even when the sequence length is held constant, methods with fewer parameters often require more memory. This is due to the "activation density" of the specific adapter. Methods that insert many small adapters throughout the transformer's depth (like IA3) create more "gradient-required" nodes in the computational graph than methods that concentrate updates (like Prefix Tuning), which often under perform.

\subsection{Memory Decomposition}
\label{app:memory}
Section 2 shows memory decomposition of a standard foundation model. The various terms in the peak memory scales as $M_{params}=\mathcal{O}(|\theta|)$ and $M_{grads}=M_{opt}=\mathcal{O}(|\theta_{trainable}|)$. However activation memory is beyond parameter count. Across layers, hidden-state activations scale at least as $M_{acts}=\mathcal{O}(BSHL)$. Self-attention introduces additional intermediate tensors. In standard implementations, attention logits scale as $\mathcal{O}(BS^2)$ per head \cite{attentionneed}. Memory-efficient implementations (e.g., recomputation-based\cite{checkpointing} or fused attention kernels\cite{flashattention}) reduce this overhead but do not eliminate the need to retain layer-wise hidden states for gradient computation. Thus, regardless of attention implementation, activation memory scales at least linearly with $\mathcal{O}(BSHL)$. Generally, activation memory frequently dominates peak memory usage.

The memory decomposition in Figure \ref{fig:memory_decomposition} and Table \ref{mem_decomp_table} were obtained by explicitly measuring the major runtime memory components during training: model weights, gradients, optimizer states, forward-pass activations, and peak training memory. Model parameter memory was computed directly from the total number of parameters and their datatype - float32. While gradient memory was similarly estimated over all trainable parameters. Optimizer-state memory was measured from the internal AdamW optimizer tensors (e.g., momentum and variance buffers) by iterating through the optimizer state dictionary and summing the size of all stored tensors. Peak GPU memory during training was measured using \textit{torch.cuda.max\_memory\_allocated()} after resetting CUDA memory statistics prior to each optimization step. We have split adapter activations into (1) adapter-specific activations from the adapter internals which LARS specifically attend to and reduce from $O(BSRL)$ to $O(BRL)$ and (2) adapter buffer, which includes activations of adapter output of shape $[BSHL]$. While adapter-specific activations remain constant at 0.035 GB as $S$ scales from 512 to 5120, adapter buffer, which is not targeted by LARS, scales with $S$.

To isolate activation-related memory, we measured both forward-pass memory and total training memory. Forward-pass memory was estimated by executing inference under \textit{torch.no\_grad()} and recording the peak allocated memory, providing a proxy for runtime activations independent of backward graph retention. Additionally, hook-based activation estimates were computed using forward hooks registered across all modules, where the memory footprint of intermediate outputs was accumulated as:
$$
M_{\text{acts}} = \sum_i \texttt{numel}(a_i)\times \texttt{element\_size}(a_i),
$$
where $a_i$ denotes the intermediate activation tensors. The adapter-specific contribution shown in Figure~\ref{fig:memory_decomposition} was then estimated by comparing the difference in peak and activation-related memory between the frozen base model and the corresponding PEFT-adapted model under identical batch size and sequence-length settings.



\begin{table*}[t]
\centering
\caption{Memory decomposition of LoRA and LARS at $S=512$ and $S=5120$. The adapter activations have been split into adapter-specific activations or internals and adapter buffer, which includes the activations of adapter outputs stored for backpropagation. }
\label{mem_decomp_table}
\resizebox{2\columnwidth}{!}{
\begin{tabular}{ccccccccc}
\hline
\textbf{}                &                              & \textbf{\begin{tabular}[c]{@{}c@{}}Base Model \\ Weights (GB)\end{tabular}} & \textbf{\begin{tabular}[c]{@{}c@{}}Base Model Est. \\ Activations (GB)\end{tabular}} & \textbf{\begin{tabular}[c]{@{}c@{}}Adapter Est.\\ Activations (GB)\end{tabular}} & \textbf{\begin{tabular}[c]{@{}c@{}}Adapter\\ Weights (MB)\end{tabular}} & \textbf{\begin{tabular}[c]{@{}c@{}}Adapter\\ Grad.(MB)\end{tabular}} & \textbf{\begin{tabular}[c]{@{}c@{}}Adapter Opt. \\ States (MB)\end{tabular}} & \textbf{\begin{tabular}[c]{@{}c@{}}Adapter\\ Buffer (GB)\end{tabular}} \\ \hline
                         & \cellcolor[HTML]{EFEFEF}LARS & \cellcolor[HTML]{EFEFEF}4.94                                                & \cellcolor[HTML]{EFEFEF}1.38                                                         & \cellcolor[HTML]{EFEFEF}0.035                                                    & \cellcolor[HTML]{EFEFEF}33.38                                           & \cellcolor[HTML]{EFEFEF}33.38                                        & \cellcolor[HTML]{EFEFEF}67.76                                                & \cellcolor[HTML]{EFEFEF}0.78                                           \\
\multirow{-2}{*}{S=512}  & LoRA                         & 4.94                                                                        & 1.38                                                                                 & 1.39                                                                             & 22.58                                                                   & 22.58                                                                & 45.16                                                                        & 0.83                                                                   \\
                         & \cellcolor[HTML]{EFEFEF}LARS & \cellcolor[HTML]{EFEFEF}4.94                                                & \cellcolor[HTML]{EFEFEF}13.84                                                        & \cellcolor[HTML]{EFEFEF}0.035                                                    & \cellcolor[HTML]{EFEFEF}33.38                                           & \cellcolor[HTML]{EFEFEF}33.38                                        & \cellcolor[HTML]{EFEFEF}67.76                                                & \cellcolor[HTML]{EFEFEF}7.8                                            \\
\multirow{-2}{*}{S=5120} & LoRA                         & 4.94                                                                        & 13.84                                                                                & 13.9                                                                             & 22.58                                                                   & 22.58                                                                & 45.16                                                                        & 8.3                                                                    \\ \hline
\end{tabular}}
\end{table*}
\begin{table*}[t]
\centering
\caption{Memory decomposition of LoRA and LARS at $S=512$ and $S=5120$ with gradient checkpointing. The reported estimated adapter activations include both adapter internals and buffers as they are significantly smaller in comparison to Table \ref{mem_decomp_table} due to gradient checkpointing. }
\label{mem_decomp_table_GC}
\resizebox{2\columnwidth}{!}{
\begin{tabular}{cccccccc}
\hline
\textbf{}                &                              & \textbf{\begin{tabular}[c]{@{}c@{}}Base Model \\ Weights (GB)\end{tabular}} & \textbf{\begin{tabular}[c]{@{}c@{}}Base Model Est. \\ Activations (GB)\end{tabular}} & \textbf{\begin{tabular}[c]{@{}c@{}}Adapter Est.\\ Activations (GB)\end{tabular}} & \textbf{\begin{tabular}[c]{@{}c@{}}Adapter\\ Weights (MB)\end{tabular}} & \textbf{\begin{tabular}[c]{@{}c@{}}Adapter\\ Grad.(MB)\end{tabular}} & \textbf{\begin{tabular}[c]{@{}c@{}}Adapter Opt. \\ States (MB)\end{tabular}} \\ \hline
                         & \cellcolor[HTML]{EFEFEF}LARS & \cellcolor[HTML]{EFEFEF}4.94                                                & \cellcolor[HTML]{EFEFEF}0.08                                                         & \cellcolor[HTML]{EFEFEF}0.05                                                     & \cellcolor[HTML]{EFEFEF}33.38                                           & \cellcolor[HTML]{EFEFEF}33.38                                        & \cellcolor[HTML]{EFEFEF}67.76                                                \\
\multirow{-2}{*}{S=512}  & LoRA                         & 4.94                                                                        & 0.08                                                                                 & 0.13                                                                             & 22.58                                                                   & 22.58                                                                & 45.16                                                                        \\
                         & \cellcolor[HTML]{EFEFEF}LARS & \cellcolor[HTML]{EFEFEF}4.94                                                & \cellcolor[HTML]{EFEFEF}0.78                                                         & \cellcolor[HTML]{EFEFEF}0.48                                                     & \cellcolor[HTML]{EFEFEF}33.38                                           & \cellcolor[HTML]{EFEFEF}33.38                                        & \cellcolor[HTML]{EFEFEF}67.76                                                \\
\multirow{-2}{*}{S=5120} & LoRA                         & 4.94                                                                        & 0.78                                                                                 & 1.33                                                                             & 22.58                                                                   & 22.58                                                                & 45.16                                                                        \\ \hline
\end{tabular}}
\end{table*}

\subsection{Effect of Gradient Checkpointing, FlashAttention, and KV Caching} 
\label{cp}
Gradient checkpointing \cite{checkpointing} reduces memory by discarding intermediate activations during the forward pass and recomputing them during backward propagation, lowering the constant factor in $M_{acts}$ but not its scaling with $B,S,H,L$. As shown in Figure \ref{fig:s_dependence} (Right), even with GC enabled, the slope 
 remains identical for all token-level PEFT methods. GC trades compute (FLOPs) for memory, but it does not change the asymptotic dependence on sequence length. LARS, by contrast, reduces the rank of the tensors that must be stored or recomputed, providing a lower growth rate that compounds with the benefits of GC. The same can be observed in Table \ref{mem_decomp_table_GC}; while the activation memory consumption is lower than that without checkpointing, the reduction still scales with respect to $S$ for LARS.

Flash-style memory-efficient attention kernels\cite{flashattention} further shrink attention-specific buffers (e.g., logits and softmax intermediates) and optimizes the $\mathcal{O}(S^2)$memory requirement of the attention matrix by computing it in tiles. While FlashAttention solves the memory spike of the attention weights, ($\mathbb{R}^{S\times S}$), it does nothing for the memory footprint of the hidden states ($\mathbb{R}^{B\times S\times R}$). LARS specifically targets these $\mathcal{O}(BSR)$ tensors, which FlashAttention leaves untouched.

 During training, KV caching \cite{kvcache} primarily serves to accelerate autoregressive decoding and does not remove the need to store or reconstruct hidden states needed for backpropagation, offering limited relief for peak activation memory in the full-sequence training regime. Consequently, while checkpointing, optimized attention, and KV caching can substantially reduce constants or improve runtime, they do not alter the asymptotic dependence of training memory on token-level activations and therefore do not eliminate the activation bottleneck.

 \subsection{Constants vs. Growth Rates}
\label{cieling}
Standard optimizations like GC \cite{checkpointing} and FlashAttention \cite{flashattention} target the constant factors of $M_{acts}$. While GC lowers the absolute memory baseline (Figure \ref{fig:s_dependence} Right), the growth rate (the slope $\frac{\partial M}{\partial S}$) remains identical for LoRA and IA3. This is because they must still materialize the full activation tensors for the backward pass.

In contrast, LARS reduces this slope by operating in a sequence-pooled subspace. By decoupling the gradient-related activations from $S$, LARS provides a superior Pareto frontier for long-context adaptation. As $S$ increases, LARS effectively raises the "Sequence Length Ceiling" for memory-constrained hardware, enabling adaptation where traditional PEFT methods would trigger Out-of-Memory (OOM) errors.

 \subsection{Context-Aware Learned Pooling}
 While fixed pooling is optimal for memory-constrained environments, uniform averaging may dilute fine-grained nuances in complex documents. To recover this expressivity, we introduce a learnable linear projection $w_{pool} \in \mathbb{R}^H$ that computes per-token importance scores. The pooled representation is calculated as:
\begin{equation}
\begin{aligned}
   x_{\text{pool}} &= \sum_{i=1}^S \alpha_i X_i \\
 \text{where,} \quad   \alpha_i &= \operatorname{Softmax}(\operatorname{pool\_proj}(X_i)) 
\end{aligned}
\end{equation}

This strategy allows the model to prioritize critical semantic anchors while ignoring syntactic markers or padding, aligning with findings that principled pooling is essential for maintaining Transformer capacity \cite{pooling}. In our evaluations (see Table 1), this variant—denoted as LARS-LP—consistently bridges the accuracy gap on challenging reasoning tasks like MMLU-Pro, offering a tunable Pareto frontier between peak memory efficiency and predictive performance.

\section{Method Details}
\label{code}
\begin{figure*}[t]
    \centering
    \includegraphics[width=1\linewidth]{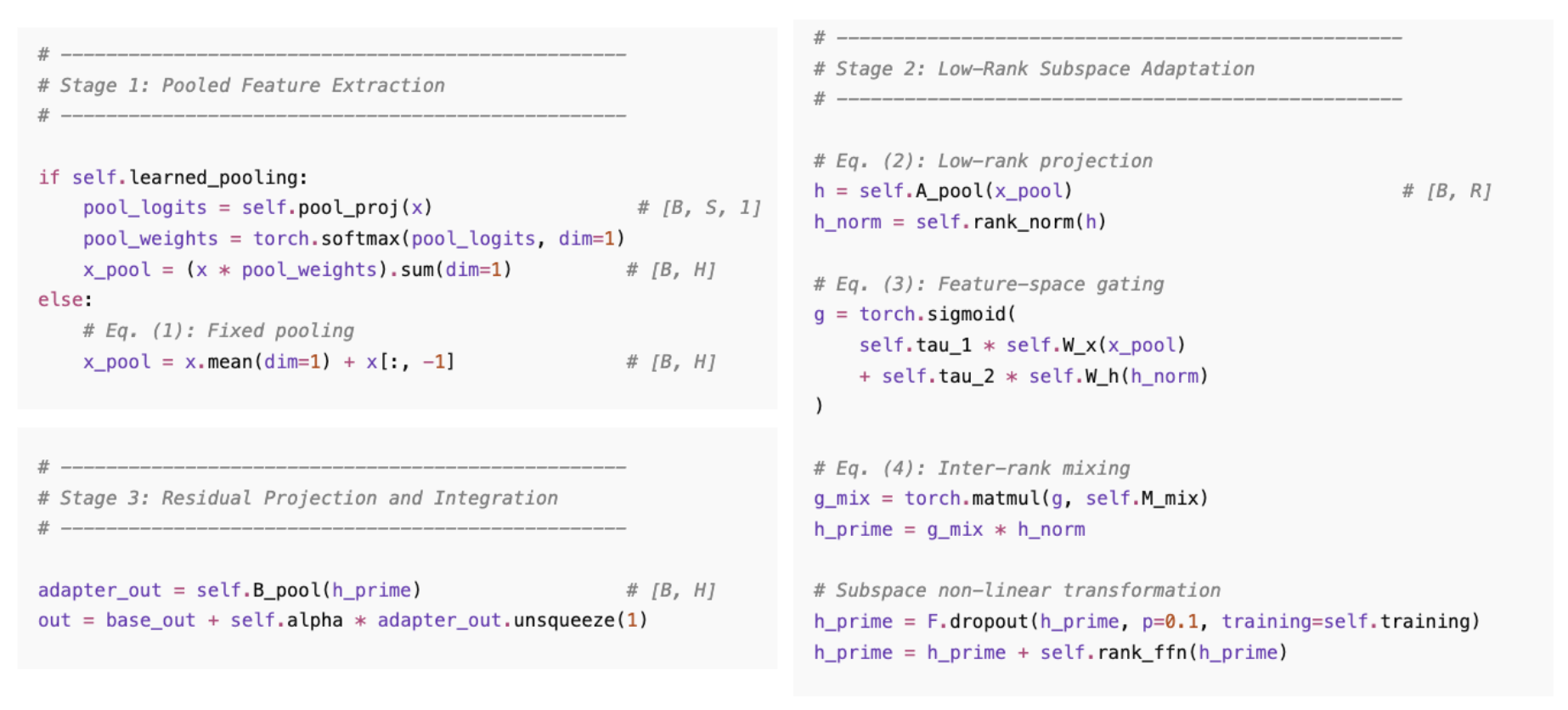}
    \caption{Code snippets corresponding to the 3 stages in LARS - Sequence pooling, low-rank subspace modulations, and residual projection via broadcasting.}
    \label{fig:code}
\end{figure*}

This section explains in the detail the implementation specifics and design choices of LARS. In addition to this, the repository will be made public on acceptance.

\subsection{Initialization and Stability Design}
While the paper focuses on the architectural formulation of LARS, the implementation introduces several stability-oriented initialization choices. The residual output projection $B_{pool}$ is initialized to zeros, ensuring that the adapter initially behaves as an identity perturbation over the frozen backbone. Similarly, the low-rank subspace gating projections $W_x$ and $W_h$ are initialized to zeros so that early training begins from a neutral gating state before progressively learning rank-space modulation patterns. In contrast, the low-rank projection $A_{pool}$ is initialized with a small Gaussian distribution (std = $0.01$) to preserve low-magnitude feature projections during the initial optimization phase. These initialization strategies improve optimization stability and prevent large early-stage perturbations to the pretrained model representations.

\subsection{Rank-Space Pipeline}
The implementation of LARS realizes the proposed low-rank subspace adaptation entirely within the compressed rank manifold of dimensionality $R$. After sequence pooling as in Stage 1 of Figure \ref{fig:code}, the pooled representation is projected into rank space and normalized using LayerNorm before gating. The gated features are then processed through three lightweight operations: (i) inter-rank mixing using the learnable matrix M\_{mix, (ii) multiplicative feature modulation (g\_mix * h\_norm), and (iii) a residual feed-forward transformation with GeLU activations. Importantly, all non-linear processing occurs in the low-dimensional subspace rather than the original hidden dimension $H$, preserving the activation-efficient design objective of LARS while improving representational expressivity.

\subsection{Residual Broadcasting}
A key implementation detail is how the adapter output is reintegrated into the token sequence shown in Stage 3 of Figure \ref{fig:code}. The adapter produces a single sequence-level residual vector of shape $[B, H]$, which is broadcast across the sequence dimension using \textit{unsqueeze(1)} before addition to the frozen backbone activations. This avoids constructing token-level adapter activations of shape [B, S, R] during the forward and backward passes. Consequently, the trainable adapter intermediates remain independent of sequence length, enabling the reduction of adapter activation complexity from $\mathcal{O}(BSRL)$ to $\mathcal{O}(BRL)$ as discussed.

\section{Additional Setup Details}
\label{App:setup}
\subsection{Tasks and Datasets}
To rigorously evaluate the memory-to-accuracy trade-off of LARS (Low-memory Activation-Rank Subspace), we utilize a diverse evaluation suite categorized into three primary domains. These tasks are designed to stress-test the model's ability to maintain high-resolution signals despite the sequence-pooling mechanism.

\textbf{1. Commonsense Reasoning} We evaluate reasoning capabilities using five core benchmarks from the LLM-adapters family. After fine-tuning, these tasks require the model to perform logical inference over everyday scenarios.
\begin{itemize}
    \item BoolQ: A reading comprehension dataset of 15,942 naturally occurring yes/no questions
\item PIQA (Physical Interaction QA): Tests the model's understanding of physical objects and their interactions 
\item SIQA (Social Interaction QA): Focuses on reasoning about social interactions and social commonsense
\item HellaSwag: A dataset that challenges the model to complete sentences by predicting the most plausible continuation of a scene
\item ARC-c (AI2 Reasoning Challenge): A "Challenge" set consisting of difficult, grade-school science questions that require more than simple retrieval
\end{itemize}

\textbf{2. General Understanding (MMLU-Pro)}
\begin{itemize}
    \item Subjects: Economics, Biology, Physics, Health, and Math. These subjects are chosen at random.
    \item Task: These subjects measure the model's ability to maintain competitive accuracy
\end{itemize}

\textbf{3. Long-Context \& Retrieval Analysis}
A critical component of our evaluation is the "Sequence Length Ceiling" test, where we evaluate if LARS can handle long inputs without the linear memory growth typical of LoRA or IA3.
\begin{itemize}
    \item QuALITY: A multiple-choice QA dataset featuring long input texts that require deep reasoning
    \item RACE: A large-scale reading comprehension dataset derived from middle and high school English exams
    \item Nanotron NIAH (Needle-In-A-Haystack): We perform passkey retrieval tasks across context lengths of 1024, 16k, and 32k tokens. This validates that LARS’s pooling mechanism does not cause "catastrophic information loss" and can retrieve localized, high-resolution signals with near-perfect accuracy
\end{itemize}

\textbf{3. Token-Level Fidelity}
To evaluate whether LARS generalizes beyond classification and retrieval tasks, we additionally benchmark it on structured prediction, span extraction, and generative summarization workloads.
\begin{itemize}
\item CoNLL-2003: A sequence labeling benchmark for named entity recognition (NER), where models must preserve token-level contextual representations across long token sequences. This evaluates whether LARS’s pooled adaptation mechanism can maintain fine-grained token discrimination required for accurate labeling.
\item SQuAD: An extractive question answering benchmark requiring precise span localization within a context passage. This tests whether LARS can retain localized contextual information necessary for accurate answer boundary prediction despite compressed adaptation activations.

\item SAMSum: A dialogue summarization benchmark focused on conversational summarization. This evaluates whether LARS can preserve long-range conversational dependencies and salient contextual information needed for coherent abstractive summaries while maintaining memory efficiency.
\end{itemize}

\subsection{Hyperparameters and Fine-Tuning Setup}
\paragraph{Models } We evaluate two base models - Llama 3.2 1B and Qwen2.5 7B Instruct. The Llama model is used for most experiments, while Qwen serves as a larger model comparison. For classification tasks we use sequence classification heads, whereas long-context, reasoning, and token-level fidelity benchmarks are trained using causal language modeling. These models are open access models available on Huggingface.

\paragraph{Hyperparameters } All models are fine-tuned for 1500 steps using the AdamW optimizer with weight decay 0.01, Cosine decay with 100 warmup steps, and gradient clipping 1.0. Other baseline-specific hyperparameters like LoRA $\alpha$, LoRA dropout, AdaLoRA init rank etc were chosen as the default values provided by HuggingFace PEFT Library. Training uses dynamic token counting to monitor throughput and GPU utilization. Other fine-tuning dataset specific hyperparameters are as listed below:
\begin{itemize}
    \item BoolQ: batch size (BS) = 8, accumulation steps (AS) = 4
    \item PIQA: BS = 8, AS = 4
    \item SIQA: BS = 8, AS = 4
    \item HellaSwag: BS = 2, AS = 16
    \item ARC-c: BS = 8, AS = 4
    \item MMLU-Pro: BS = 2, AS = 8
    \item QuALITY: BS = 2, AS = 16
    \item RACE: BS = 2, AS = 16
    \item Nanotron NIAH: BS = 1
\end{itemize}

The above dataset-specific BS and AS were chosen specifically to run the experiments on at max a single GPU of L40S for Llama or H200 for Qwen models.

Learning rates for each baseline including LARS were obtained via tuning on BoolQ and the same were used for all experiments as the primary contribution of this work is memory reduction with comparable, which is determined by architectural design and is invariant to hyperparameter choices. While tuning can influence accuracy, it does not affect the observed memory reductions. To ensure that this choice do not adversarially affect the baselines, we conducted learning rate sweeps (5 values) for LARS, LoRA, AdaLoRA, and IA3 on SIQA and HellaSwag as shown in Table \ref{lrsweep}. The results show that the learning rates used in the paper (italics) achieve performance close to the best (bold) for each method, with only minor deviations.

While variance across random seeds can influence downstream accuracy metrics, the primary contribution of this work is reducing the sequence-length scaling of adapter-specific activation memory, which is determined by the architectural structure of the method and remains invariant to initialization or seed choice. Consequently, our evaluation prioritizes consistent memory, throughput, and activation-scaling comparisons across methods and hardware settings. We nevertheless include additional experiments across multiple datasets, tasks, and model scales to demonstrate that the observed performance trends are stable and not isolated to a single benchmark configuration. Additionally, we recommend careful tuning for applications where accuracy gains are of extreme importance.

\begin{table*}[t]
\centering
\caption{Learning rate sweep for LARS, LoRA, AdaLoRA, and IA3 on SIQA and HellaSwag. }
\label{lrsweep}
\resizebox{2\columnwidth}{!}{

\begin{tabular}{ccccccccccccc}
\hline
\rowcolor[HTML]{EFEFEF} 
\textbf{}   &  & \multicolumn{5}{c}{\cellcolor[HTML]{EFEFEF}\textbf{SIQA}}                       & \textbf{} & \multicolumn{5}{c}{\cellcolor[HTML]{EFEFEF}\textbf{Hellaswag}}                                    \\ \hline
\textbf{lr} &  & \textbf{1e-2} & \textbf{1e-3}  & \textbf{1e-4}  & \textbf{1e-5} & \textbf{1e-6} & \textbf{} & \textbf{1e-2} & \textbf{1e-3}           & \textbf{1e-4}           & \textbf{1e-5} & \textbf{1e-6} \\ \hline
\rowcolor[HTML]{EFEFEF} 
LARS        &  & 76.61         & \textbf{78.85} & \textit{78.04} & 78.75         & 70.21         &           & 25            & 85.62                   & \textit{\textbf{90.2}}  & 90.2          & 75.62         \\
LoRA        &  & 42.94         & 76.25          & \textit{78.77} & \textbf{79.2} & 75            &           & 30            & 70.11                   & \textit{\textbf{90.45}} & 90.21         & 75.62         \\
\rowcolor[HTML]{EFEFEF} 
IA3         &  & 71.25         & \textit{73.75} & \textbf{74.92} & 48.75         & 45            &           & 85.62         & \textit{\textbf{85.77}} & 85.26                   & 65.2          & 45            \\
AdaLoRA     &  & 71.25         & \textbf{78.5}  & 76.61          & 63.75         & 25            &           & 30            & \textit{\textbf{90.21}} & 85.85                   & 75.2          & 25            \\ \hline
\end{tabular}}
\end{table*}

\paragraph{Hardware}
For our experiments we had access to 4 hardwares:
\begin{itemize}
    \item NVIDIA L40S: All Llama 1B experiments were run on this. GPU memory available was 45GB.
    \item NVIDIA H200: All Qwen 7B and model scaling experiments were run on this. GPU memory available was 145GB.
    \item AMD EPYC: This powerful CPU was used only for Table 2
    \item Raspberry Pi 5: This 8GB edge device was used only validation
\end{itemize}

\paragraph{Memory, Throughput, and Power Measurement Methodology}
To accurately measure memory efficiency and model throughput, we implemented the following procedure:
\begin{itemize}
    \item GPU Peak Memory: Before training, GPU memory statistics are reset with \textit{torch.cuda.reset\_peak\_memory\_stats()} and \textit{torch.cuda.empty\_cache()}. During training, peak memory usage is logged after each optimizer step using \textit{torch.cuda.max\_memory\_allocated()}.
    \item CPU Peak Memory: Peak CPU memory is tracked using Python’s \textit{psutil} library - \textit{process = psutil.Process(); cpu\_mem\_mb = process.memory\_info().rss / 1e6}. The maximum of this provides the peak memory usage.
    \item Throughput Measurement: Token throughput is computed dynamically as
    $$ \text{tokens/sec} = \frac{\text{total tokens processed per AS}}{\text{elapsed wall-clock time in sec.}}
    $$
    For inference throughput, we consider the total tokens in the evaluation set and total time taken for evaluation.

    \item Power Measurement: For CPU experiments, average power consumption is estimated using real-time CPU utilization from the \textit{psutil} library together with a linear power model:
$$
P_{\text{cpu}} = P_{\text{idle}} + u \cdot (P_{\text{max}} - P_{\text{idle}})
$$
where $u \in [0,1]$ denotes CPU utilization, $P_{\text{idle}}$ is the idle CPU power, and $P_{\text{max}}$ is the maximum processor power draw.

\item Energy Measurement: Energy consumption per optimization step is computed as:
$$
E_{\text{step}} = P_{\text{cpu}} \times t_{\text{step}}
$$
where $P_{\text{cpu}}$ is the average power consumption during the step and $t_{\text{step}}$ is the elapsed wall-clock time for that optimization step.
\end{itemize}

On the resource-constrained Raspberry Pi 5 (8GB), LARS achieved a memory footprint of 3.008 GB and a throughput of 3.20 tokens/sec, demonstrating performance parity with established methods while maintaining a lower memory profile than LoRA and IA3. This efficiency scales effectively to more performant CPUs. On the AMD EPYC 9474F, LARS consistently outperformed LoRA in memory efficiency. Notably, at a sequence length of 1024, LARS showed nearly $2\times$ reduction—while maintaining competitive throughput. \textit{These results confirm that LARS is a highly viable candidate for deploying large-scale activations on CPU-only environments, including edge devices.}

 \section{Additional Results}
\label{App:ablations}
\subsection{Long Context Results}

\begin{figure}[t]
    \centering
    \includegraphics[width=1\linewidth]{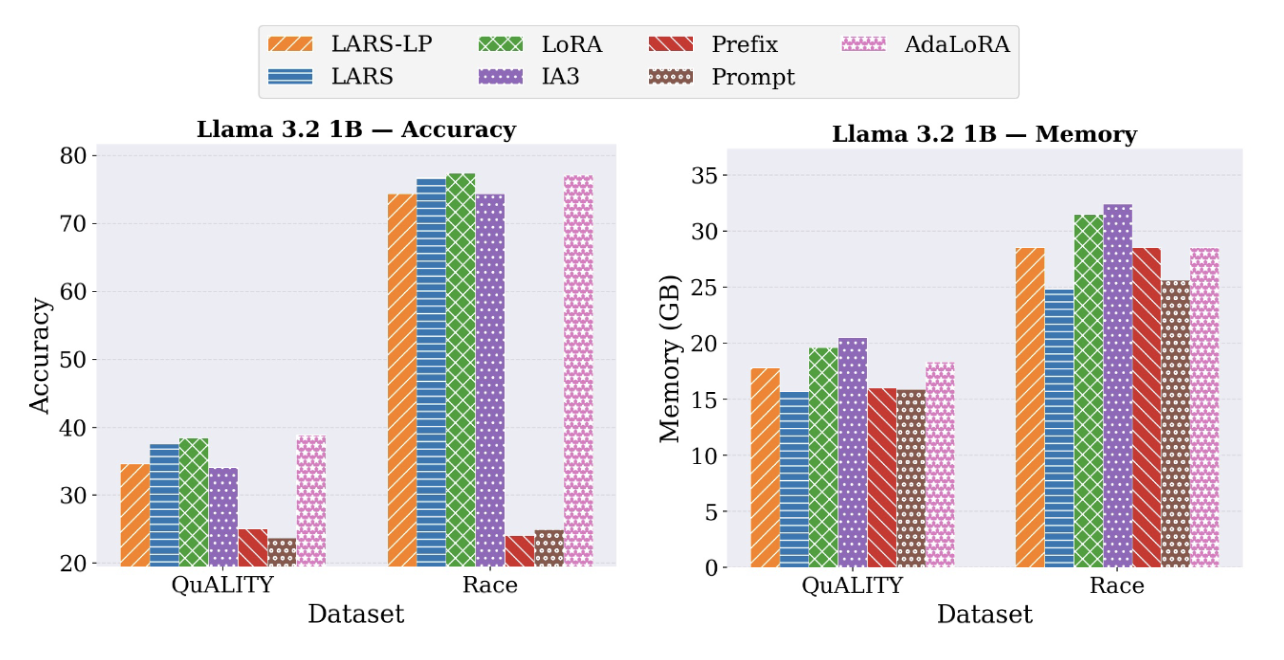}
    \caption{Comparison of accuracy and memory on long context tasks using QuALITY and Race datasets for models Llama 3.2 1B model.}
    \label{fig::long-context2}
\end{figure}

Figures \ref{fig:long-context1} and \ref{fig::long-context2} illustrate the memory consumption and predictive accuracy across two long-context benchmarks, QuALITY and Race, using Qwen 7B models and Llama 3.2 1B respectively. We compare our proposed LARS and LARS-LP methods against standard PEFT baselines, including LoRA, IA3, Prefix Tuning, Prompt Tuning, and AdaLoRA. Across both model scales and datasets, LARS consistently demonstrates superior memory efficiency relative to all other high-performing adapters. In terms of accuracy, LARS and LARS-LP achieve performance parity with LoRA and AdaLoRA, despite their reduced memory requirements across both datasets and models. Collectively, these results indicate that \textit{LARS provides a favorable trade-off for long-context applications, offering the high-rank representation power of LoRA with a memory profile more closely resembling (or bettering) more restrictive parameter-efficient methods.}

 \subsection{Latency of LARS}
 \begin{figure}[h]
\centering
\includegraphics[width=\columnwidth]{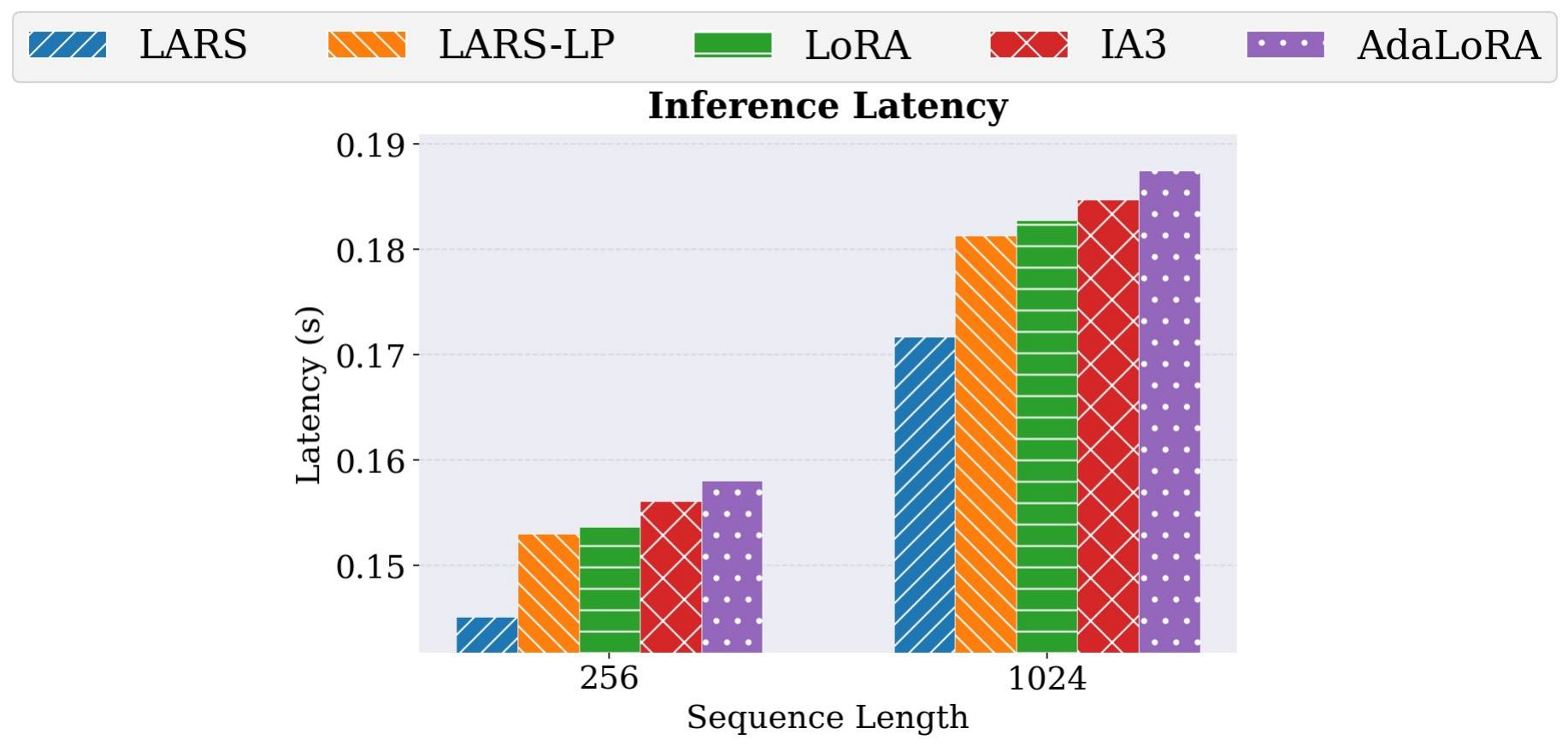}
\caption{Inference latency of LARS and other baselines.}
\label{fig:latency}
\end{figure}

Figure \ref{fig:latency} follows the inference throughput shown in Figure \ref{fig:throughput}. The results demonstrate that LARS consistently achieves the lowest inference latency (the fastest processing time) at both sequence lengths of 256 and 1024. As expected, latency increases for all methods as the sequence length grows, but LARS maintains its performance advantage, while AdaLoRA consistently exhibits the highest latency. This indicates that LARS is not only memory-efficient, but also faster in terms of inference speed.

\subsection{Impact of Target Modules}
\begin{figure}[h]
\centering
\includegraphics[width=\columnwidth]{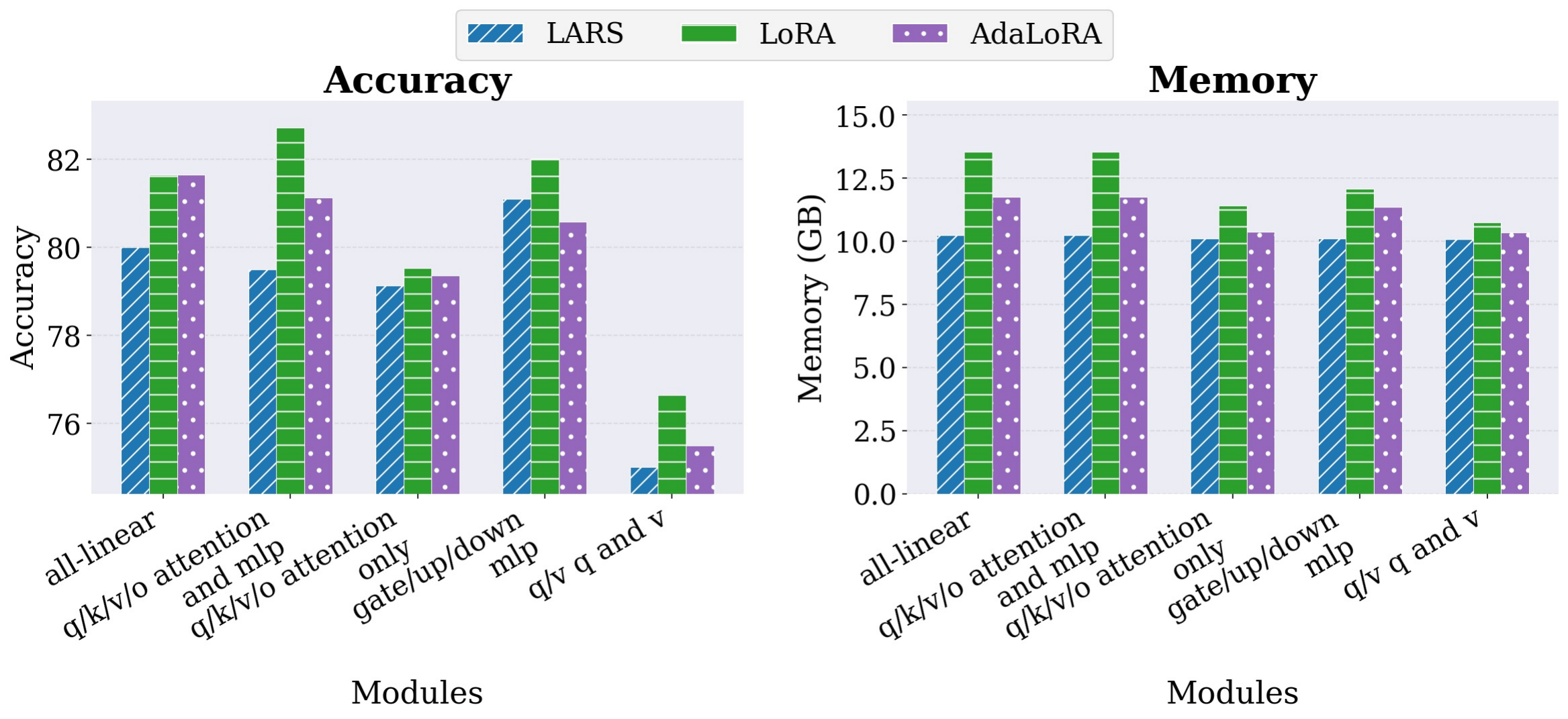}
\caption{Comparison of Memory Usage (GB) and Accuracy across different target modules for the LARS, LoRA, and IA3 fine-tuning methods. }
\label{fig:target_modules}
\end{figure}
Figure \ref{fig:target_modules} evaluates how targeting different architectural modules—such as attention layers, MLP components, or all linear layers—affects the performance and resource usage of LARS, LoRA, and AdaLoRA. The accuracy plot on the left shows that while LoRA often achieves the highest overall accuracy, LARS remains a strong competitor, particularly when targeting the "gate/up/down mlp" modules. On the right, the memory plot highlights LARS's primary advantage: its GPU footprint remains remarkably stable and low (around 10 GB) regardless of which modules are modified. In contrast, LoRA and AdaLoRA exhibit much higher memory demands and greater sensitivity to the specific modules being tuned, making LARS the more predictable choice for hardware-constrained environments..

\subsection{Effect of Rank}
\begin{figure}[h]
\centering
\includegraphics[width=\columnwidth]{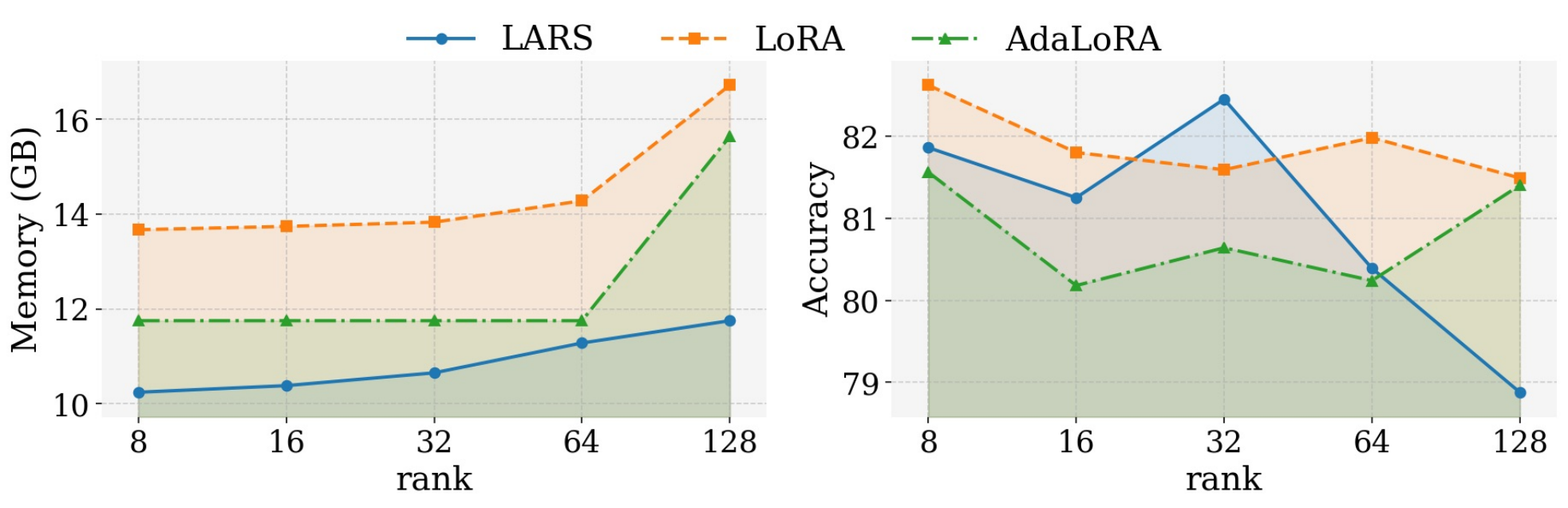}
\caption{Comparison of Memory Usage (GB) and Accuracy for the LARS, LoRA, and AdaLoRA methods across different ranks (r)}
\label{fig:rank}
\end{figure}

Figure \ref{fig:rank} compares the trade-offs between hardware resource efficiency and model performance for three fine-tuning methods: LARS, LoRA, and AdaLoRA. The left plot illustrates that LARS maintains a consistently low and stable GPU memory footprint even as the rank increases. In contrast, both LoRA and AdaLoRA require significantly more memory, which scales up to nearly 17 GB at higher ranks. However, the accuracy plot on the right reveals a performance trade-off; while LoRA and AdaLoRA maintain relatively stable accuracy across all rank values, LARS is highly sensitive to the rank configuration. It reaches a competitive peak at $r=32$ but suffers a sharp decline in accuracy after $r=128$. Additionally, LARS demonstrates superior inference speed across all tested ranks as shown in Appendix \ref{rank_throughput}. These results suggest that while \textit{LARS is ideal for memory-constrained environments}, it requires more precise hyperparameter tuning than its counterparts. 

\subsubsection{Throughput with Increasing Rank}
\label{rank_throughput}
 \begin{figure}[h]
\centering
\includegraphics[width=\columnwidth]{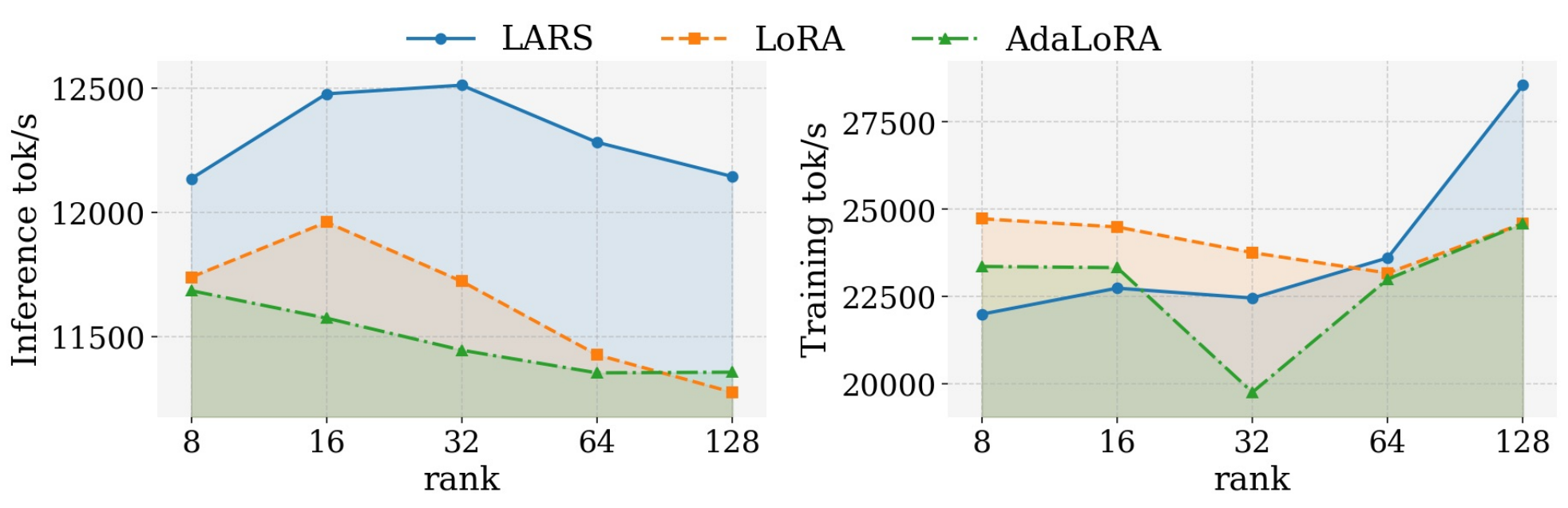}
\caption{Inference and training throughput of LARS and baselines with increasing ranks.}
\label{fig:rank_throughput}
\end{figure}

Figure \ref{fig:rank_throughput} adds onto Figure \ref{fig:rank}, from a throughput efficiency measured in tokens per second. The left plot shows that LARS consistently achieves higher inference throughput than the baselines, maintaining a clear speed advantage even as the rank increases. On the right, while LARS begins with slightly lower training throughput at smaller ranks, it experiences a dramatic performance surge at $r=128$, significantly outperforming both LoRA and AdaLoRA, which tend to slow down as rank complexity grows.

\subsection{Effect of Dataset Size}
\begin{figure}[h]
\centering
\includegraphics[width=1\columnwidth]{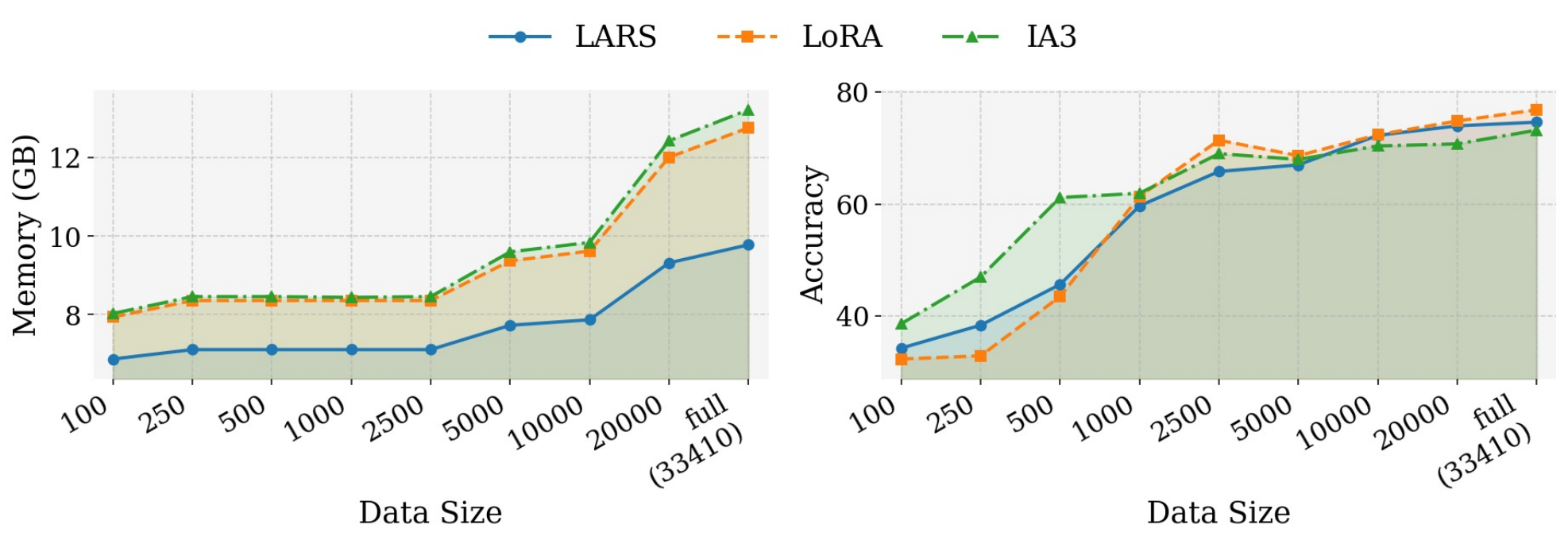}
\caption{Comparison of Memory Usage (GB) and Accuracy across increasing Data Sizes for the LARS, LoRA, and IA3 fine-tuning methods. }
\label{fig:data_size}
\end{figure}

Figure \ref{fig:data_size} examines how training data size affects both memory consumption and model accuracy across the LARS, LoRA, and IA3 methods. The left plot highlights that LARS is the most resource-friendly option, consistently requiring the least amount of GPU memory as the dataset scales. On the right, the accuracy plot shows that while IA3 takes an early lead in low-data scenarios, LoRA eventually reaches the highest accuracy at the "full" dataset size. \textit{LARS serves as a strong middle ground, providing competitive accuracy while maintaining a significantly smaller memory footprint} than its counterparts throughout the entire scaling process.

\subsection{Needle-In-A-Haystack (NIAH) Retrieval}
\begin{table}[h]
\centering
\caption{Accuracy of Needle-In-A-Haystack (NIAH) style retrieval of a passkey.}
\label{tab:niah}
\resizebox{0.75\columnwidth}{!}{
\begin{tabular}{ccccc}
\hline
\rowcolor[HTML]{EFEFEF} 
              & \textbf{LARS} & \textbf{LoRA} & \textbf{AdaLoRA} & \textbf{IA3} \\ \hline
\textbf{1024} & 99.2          & 99.2          & 99.2             & 99.2         \\
\rowcolor[HTML]{EFEFEF} 
\textbf{16k}  & 98.4          & 98.4          & 98.4             & 98.4         \\
\textbf{32k}  & 72.6          & 73.2          & 72.6             & 72.6         \\ \hline
\end{tabular}}
\end{table}
To evaluate whether LARS’s sequence-pooling mechanism results in catastrophic information loss, we conducted a passkey retrieval task in a Needle-In-A-Haystack style across context lengths of 1024, 16k, and 32k tokens using the Nanotron NIAH dataset \cite{nanotron}. LARS, LoRA, AdaLoRA, and IA3 achieved near-perfect accuracy at lengths up to 16k; however, performance across all methods declined at the 32k context length as shown in Table \ref{tab:niah}. These results demonstrate that LARS’s memory efficiency maintains parity with standard PEFT methods.

\begin{table*}[]
\centering
\caption{Accuracy and memory consumption of LoRA and LARS in combination with DropBP\cite{dropbp} and LoRA-FA \cite{lorafa}. These results reinstate theire complementary nature to the proposed method, LARS.}
\label{dropbpfa}
\resizebox{2\columnwidth}{!}{
\begin{tabular}{lrrrrrrrrrrrr}
\hline
\rowcolor[HTML]{EFEFEF} 
\multicolumn{1}{c}{\cellcolor[HTML]{EFEFEF}\textbf{}} & \multicolumn{2}{c}{\cellcolor[HTML]{EFEFEF}\textbf{LoRA}}   & \multicolumn{2}{c}{\cellcolor[HTML]{EFEFEF}\textbf{LARS}}   & \multicolumn{2}{c}{\cellcolor[HTML]{EFEFEF}\textbf{\begin{tabular}[c]{@{}c@{}}LoRA\\ +FA\end{tabular}}} & \multicolumn{2}{c}{\cellcolor[HTML]{EFEFEF}\textbf{\begin{tabular}[c]{@{}c@{}}LARS\\ +FA\end{tabular}}} & \multicolumn{2}{c}{\cellcolor[HTML]{EFEFEF}\textbf{\begin{tabular}[c]{@{}c@{}}LoRA+\\ DropBP\end{tabular}}} & \multicolumn{2}{c}{\cellcolor[HTML]{EFEFEF}\textbf{\begin{tabular}[c]{@{}c@{}}LARS+\\ DropBP\end{tabular}}} \\
                                                      & \multicolumn{1}{l}{acc (\%)} & \multicolumn{1}{l}{mem (GB)} & \multicolumn{1}{l}{acc (\%)} & \multicolumn{1}{l}{mem (GB)} & \multicolumn{1}{l}{acc (\%)}                       & \multicolumn{1}{l}{mem (GB)}                       & \multicolumn{1}{l}{acc (\%)}                       & \multicolumn{1}{l}{mem (GB)}                       & \multicolumn{1}{l}{acc (\%)}                         & \multicolumn{1}{l}{mem (GB)}                         & \multicolumn{1}{l}{acc (\%)}                         & \multicolumn{1}{l}{mem (GB)}                         \\ \hline
\rowcolor[HTML]{EFEFEF} 
\textbf{BoolQ}                                        & 83                           & 13.67                        & 81.1                         & 10.24                        & 81.77                                              & 10.99                                              & 81.94                                              & 9.97                                               & 70.72                                                & 11.89                                                & 74.79                                                & 9.67                                                 \\
\textbf{SIQA}                                         & 77.28                        & 20.55                        & 76.74                        & 14.45                        & 77.17                                              & 14.52                                              & 77.04                                              & 10.58                                              & 73.44                                                & 12.34                                                & 73.95                                                & 11.85                                                \\ \hline
\end{tabular}}
\end{table*}
\subsection{Scaling across Model Sizes}
\begin{figure}
    \centering
    \includegraphics[width=1\linewidth]{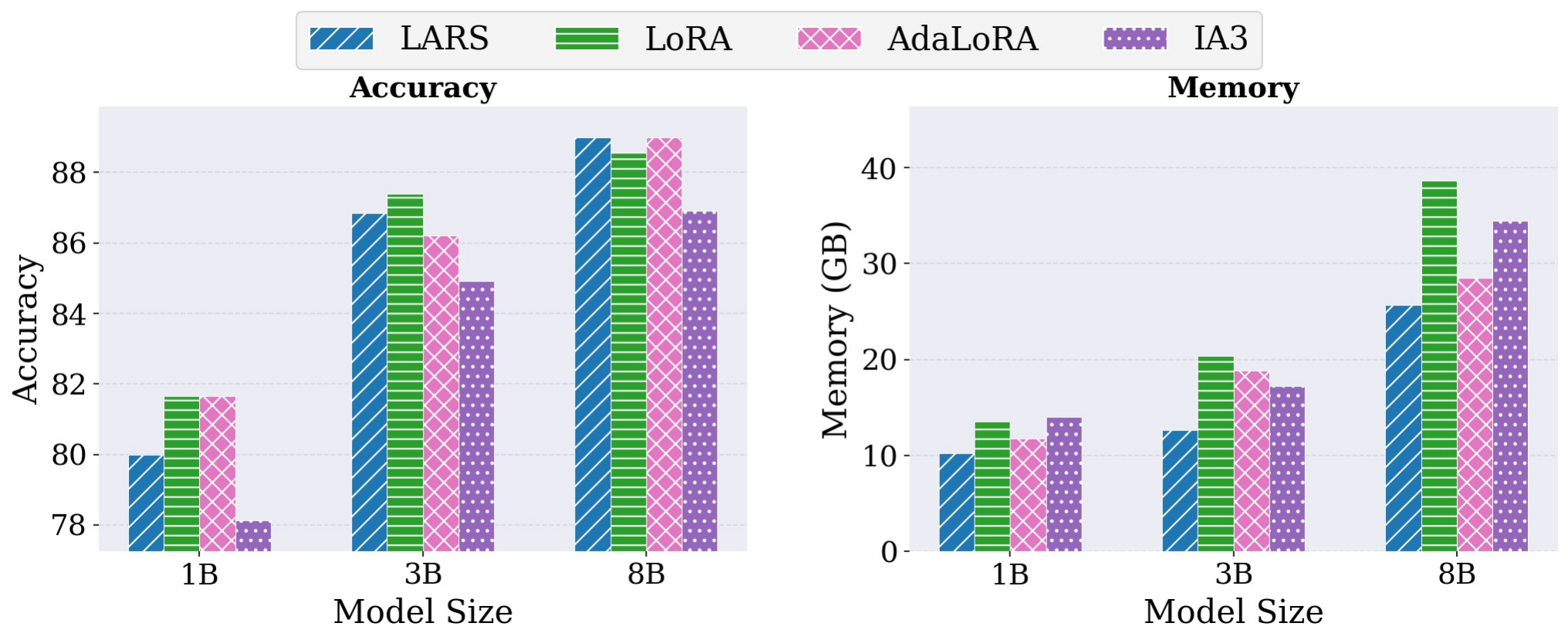}
    \caption{Impact of different model sizes on Accuracy and Memory Usage (GB) of LARS and baselines.}
    \label{fig:model_size}
\end{figure}

We further evaluate LARS on Llama models ranging from 1B to 8B parameters to examine whether its memory advantages persist across scales. As shown in Figure \ref{fig:model_size}, LARS consistently reduces peak training memory while maintaining accuracy comparable to LoRA across all model sizes. While recent work often focuses on much larger models, smaller models remain the primary target for on-device and resource-constrained adaptation, where memory and compute budgets are limited. Evaluating within this range therefore reflects the practical deployment regime for edge devices, demonstrating that LARS can lower the hardware requirements for adapting models in such settings.

\section{Complementary methods}
\label{complementary methods}

In this section we explore the complementary nature of existing widely-used memory-reduction methods including systems-level optimization methods like FlashAttention \cite{flashattention}, Gradient Checkpointing \cite{checkpointing}, as well as algorithmic methods like DropBP\cite{dropbp} and LoRA-FA\cite{lorafa}. Note that these methods are complementary, not competing, to LARS. These methods do not specifically target the sequence-length scaling of adapters and the adapter's activation memory that grows with it, which is the core contribution of LARS. Additionally, we show that LARS can be combined with these methods for additional resource reduction.

\subsection{FlashAttention}
\label{flashattention}
 \begin{figure}[h]
\centering
\includegraphics[width=\columnwidth]{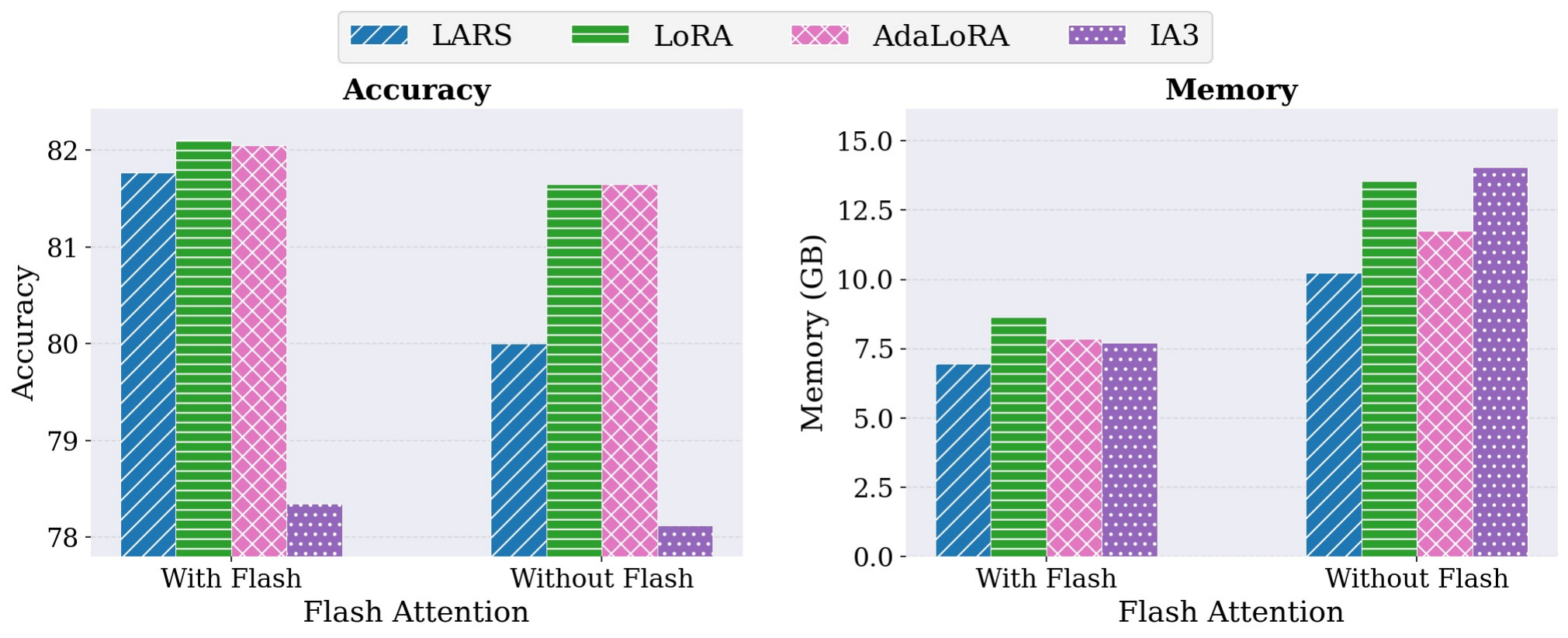}
\caption{Accuracy and Memory Consumption of LARS and baselines with and without FlashAttention.}
\label{fig:flashattentions}
\end{figure}

Figure \ref{fig:flashattentions} examines the impact of FlashAttention on accuracy and memory consumption across the LARS, LoRA, AdaLoRA, and IA3 methods. The left plot indicates that while FlashAttention consistently boosts accuracy for all methods, its impact on LARS is particularly significant, helping it bridge the performance gap with LoRA and AdaLoRA. On the right, the memory plot highlights that FlashAttention substantially reduces the GPU footprint for every technique. Notably, LARS maintains its status as the most resource-efficient option in both scenarios, requiring the least amount of memory whether FlashAttention is enabled or not.

\subsection{Gradient Checkpointing}
\label{checkpointing}

Figure \ref{fig:accuracy-memory-cp} provides a comparison of the accuracy vs. memory trade-off for several parameter-efficient fine-tuning (PEFT) methods, including LARS, LoRA, AdaLoRA, IA3, and Prompt tuning under gradient checkpointing. Compared to Figure \ref{fig:accuracy-memory}, while gradient checkpointing slightly reduced accuracy significantly saves memory for all methods, LARS still consumes relatively lower memory among all baselines while maintaining comparable performance. Similar results can be expected from methods like MEFT \cite{meft} which trades off recomputation for memory savings. Notably, LARS maintains its status as the most resource-efficient option in both scenarios, requiring the least amount of memory whether Gradient Checkpointing is enabled or not.

\subsection{Algorithmic Methods}
Besides systems-level optimization methods described above, there exists some prior work like DropBP \cite{dropbp} and LoRA-FA \cite{lorafa} that target activation memory from the algorithmic side. Specifically, DropBP drops backprop from randomly chosen pre-fixed fraction of layers to save activation memory and LoRA-FA freezes low-rank A matrix in LoRA style adapters. These methods are complementary to the asymptotic reduction of adapter activation memory and can be combined with LARS easily to achieve better efficiency. Figure \ref{dropbpfa} confirms that (i) plain LARS is capable of achieving more memory reduction than LoRA+FA[2] and LoRA+DropBP[3] and (ii) LARS can be combined with these methods to achieve more efficiency.

\end{document}